\documentclass[11pt,a4paper]{article}
\usepackage[hyperref]{emnlp-ijcnlp-2019}
\usepackage{times}
\usepackage{latexsym}
\usepackage{subfigure}
\usepackage{mathtools}
\usepackage{graphicx}
\usepackage{amsmath}
\usepackage{adjustbox}
\usepackage{caption}
\usepackage{booktabs}
\usepackage{makecell}
\usepackage{multirow}
\usepackage{amsmath,environ}
\usepackage{tabularx}
\usepackage{graphicx}
\usepackage{comment}
\usepackage{amssymb}% http://ctan.org/pkg/amssymb
\usepackage{pifont}
\usepackage[english]{babel}
\usepackage[utf8]{inputenc}
\usepackage{algorithm}
\usepackage[noend]{algpseudocode}

\usepackage{url}

\aclfinalcopy % Uncomment this line for the final submission

\title{Tree Transformer: Integrating Tree Structures into Self-Attention}

\author{Yau-Shian Wang\quad Hung-Yi Lee\quad Yun-Nung Chen \\National Taiwan University, Taipei, Taiwan\\
{\tt king6101@gmail.com\quad hungyilee@ntu.edu.tw\quad y.v.chen@ieee.org}\\}

\date{}

\begin{document}
\maketitle
\begin{abstract}
%Language is inherently hierarchical that small units gradually form larger units.
%However, Transformer is not explicitly hierarchical structured, neither interpretable.
%Pre-training Transformer form large-scale texts learns high quality language representations.
%The pre-trained Transformer can be further fine-tuned and achieves state-of-the-art results on various NLP tasks.
Pre-training Transformer from large-scale raw texts and fine-tuning on the desired task have achieved state-of-the-art results on diverse NLP tasks.
However, it is unclear what the learned attention captures.
The attention computed by attention heads seems not to match human intuitions about hierarchical structures.
This paper proposes Tree Transformer, which adds an extra constraint to attention heads of the bidirectional Transformer encoder in order to encourage the attention heads to follow tree structures.
The tree structures can be automatically induced from raw texts by our proposed ``\emph{Constituent Attention}'' module, which is simply implemented by self-attention between two adjacent words.
With the same training procedure identical to BERT, 
the experiments demonstrate the effectiveness of Tree Transformer in terms of inducing tree structures, better language modeling, and further learning more explainable attention scores\footnote{The source code is publicly available at \url{https://github.com/yaushian/Tree-Transformer}.}.
\end{abstract}

\section{Introduction}
%架構: 先說明"constituent"是什麼，parsing是個重要的問題 ->許多前作設法利用某些特殊架構來利用這個資訊->再來介紹unsupervised parsing重要性，他可以讓model更explainable->前人有用lstm recursive架構卻沒有人使用Transformer來做->我們希望能對於tranformer新增一些限制，來鼓勵Transformer可以unsupervised學習這些架構，並且讓他更explainable->概略提語其他的unsupervised parsing的方法不同處 -> 說明model的優點，以及最後的結果

%Although natural language is spoken or written in a sequential manner that one word follows another,
%language is actually constructed by attaching constituents to each other in hierarchical structures.
%The constituents, such as words, phrases or clauses, function as meaningful language units to form larger constituents with the rules designed by linguistics.
%Human inherently understands and processes language following the constituency structures.
Human languages exhibit a rich hierarchical structure which is currently not exploited nor mirrored by the self-attention mechanism that is the core of the now popular Transformer architecture.
Prior work that integrated hierarchical structure into neural networks either used recursive neural networks (Tree-RNNs)~\citep{goller,Socher:2011:PNS:3104482.3104499,tree-lstm} or simultaneously generated a syntax tree and language in RNN~\citep{dyer-rnng:16}, which have shown beneficial for many downstream tasks~\citep{aharoni-goldberg-2017-towards,eriguchi-etal-2017-learning,Strubell2018LinguisticallyInformedSF,zaremoodi-haffari-2018-incorporating}.
Considering the requirement of the annotated parse trees and the costly annotation effort, most prior work relied on the supervised syntactic parser.
However, a supervised parser may be unavailable when the language is low-resourced or the target data has different distribution from the source domain.

Therefore, the task of learning latent tree structures without human-annotated data, called \emph{grammar induction}~\citep{Carroll92twoexperiments,klein-manning-2002-generative,Smith05guidingunsupervised}, has become an important problem and attractd more attention from researchers recently.
Prior work mainly focused on inducing tree structures from recurrent neural networks~\citep{PRPN,on-lstm} or recursive neural networks~\citep{RL-parsing,diora}, while integrating tree structures into Transformer remains an unexplored direction.
%Hence, the task of learning latent tree structures without human expert annotated data becomes an important problem and attracts more attention from researchers recently.

Pre-training Transformer from large-scale raw texts successfully learns high-quality language representations.
By further fine-tuning pre-trained Transformer on desired tasks, wide range of NLP tasks obtain the state-of-the-art results~\citep{GPT-2,BERT,BERT-generation}.
However, what pre-trained Transformer self-attention heads capture remains unknown.
Although an attention can be easily explained by observing how words attend to each other, only some distinct patterns such as attending previous words or named entities can be found informative~\citep{visualize-Transformer}.
The attention matrices do not match our intuitions about hierarchical structures.

In order to make the attention learned by Transformer more interpretable and allow Transformer to comprehend language hierarchically, we propose Tree Transformer, which integrates tree structures into bidirectional Transformer encoder.
At each layer, words are constrained to attend to other words in the same constituents.
This constraint has been proven to be effective in prior work~\citep{phrase-attention}.
Different from the prior work that required a supervised parser, in Tree Transformer, the constituency tree structures is automatically induced from raw texts by our proposed ``\emph{Constituent Attention}'' module, which is simply implemented by self-attention. 
Motivated by Tree-RNNs, which compose each phrase and the sentence representation from its constituent sub-phrases, Tree Transformer gradually attaches several smaller constituents into larger ones from lower layers to higher layers.

%has to aggregate the word information hierarchically, from small constituent to large constituent, from phrase level to sentence level.
%We propose Tree Transformer, which  by multiplying self-attention probability by a constituent prior.
%The constituent prior constrains the words attend to other words within the same constituents and is automatically learned from our propose "Constituent Attention" module.

The contributions of this paper are 3-fold:
\begin{itemize}
    \item Our proposed Tree Transformer is easy to implement, which simply inserts an additional ``Constituent Attention'' module implemented by self-attention to the original Transformer encoder, and achieves good performance on the unsupervised parsing task.
    \item As the induced tree structures guide words to compose the meaning of longer phrases hierarchically, Tree Transformer improves the perplexity on masked language modeling compared to the original Transformer.
    \item The behavior of attention heads learned by Tree Transformer expresses better interpretability, because they are constrained to follow the induced tree structures.
    By visualizing the self-attention matrices, our model provides the information that better matchs the human intuition about hierarchical structures than the original Transformer. 
    %\item Following the spirit of Transformer, Tree Transformer is only built upon self-attention, where self-attention attachs constituents to each other.
    %The proposed Tree Transformer keeps the advantages of Transformer that allow to model word dependencies regardless of their distance.
    %\item By simply adding an additional module to the Transformer encoder, our work is easier to implement but the F1 scores of the induced tree structures are on par with the prior work such as PRPN~\cite{PRPN} or URNNG~\cite{urnng}.
    %\item The behavior in attention heads learned by Tree Transformer expresses better explanability, because they are constrained to follow the induced tree structures.
    %The experiments demonstrate that the attention heat maps successfully capture hierarchical structures.
    %\item The induced tree structures can serve as another source for Transformer to access positional information other than positional encoding~\cite{position-encoding}, which helps Tree Transformer achieve better performance on language modeling. %這點沒有做實驗證實，先寫上如果最後沒有實驗能證實就拿掉
\end{itemize}

% V: 我覺得可能可以把introduction裡面related work和這裡一起組成一個章節放成 Section 2: Related Work
%好的 那就這樣好了
\section{Related Work}
This section reviews the recent progress about grammar induction.
Grammar induction is the task of inducing latent tree structures from raw texts without human-annotated data.
%The models utilize the induced latent tree structures to guide the text encoding in a hierarchical order.
%To better optimize the tasks such as language modeling, the models have to induce reasonable tree structures coherent to human experts~\cite{maillard,Choi2018LearningTC}.
The models for grammar induction are usually trained on other target tasks such as language modeling.
To obtain better performance on the target tasks, the models have to induce reasonable tree structures and utilize the induced tree structures to guide text encoding in a hierarchical order.
One prior attempt formulated this problem as a reinforcement learning (RL) problem~\citep{RL-parsing}, where the unsupervised parser is an actor in RL and the parsing operations are regarded as its actions.
The actor manages to maximize total rewards, which are the performance of downstream tasks.

PRPN~\citep{PRPN} and On-LSTM~\citep{on-lstm} induce tree structures by introducing a bias to recurrent neural networks.
PRPN proposes a parsing network to compute the syntactic distance of all word pairs, and a reading network utilizes the syntactic structure to attend relevant memories.
On-LSTM allows hidden neurons to learn long-term or short-term information by the proposed new gating mechanism and new activation function.
In URNNG~\citep{urnng}, they applied amortized variational inference between a recurrent neural network grammar (RNNG)~\citep{dyer-rnng:16} decoder and a tree structures inference network, which encourages the decoder to generate reasonable tree structures.
DIORA~\citep{diora} proposed using inside-outside dynamic programming to compose latent representations from all possible binary trees.
The representations of inside and outside passes from same sentences are optimized to be close to each other.
Compound PCFG~\citep{CPCFG} achieves grammar induction by maximizing the marginal likelihood of the sentences which are generated by a probabilistic context-free grammar (PCFG) in a corpus.

\section{Tree Transformer}\label{sec:tree_Transformer}

\begin{figure*}[t!]
  \centering
    \includegraphics[width=\linewidth]{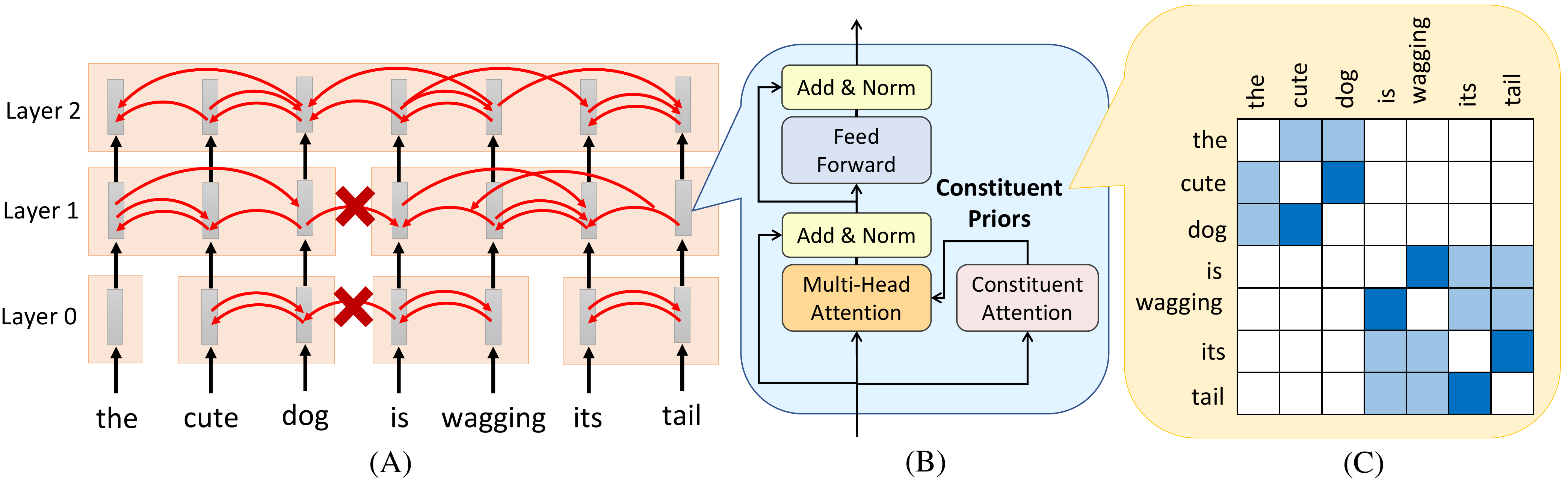}
\caption{(A) A 3-layer Tree Transformer, where the  blocks are constituents induced from the input sentence.  
The two neighboring constituents may merge together in the next layer, so the sizes of constituents gradually grow from layer to layer. 
The red arrows indicate the self-attention. 
(B) The building blocks of Tree Transformer.
(C) Constituent prior $C$ for the layer $1$.}
  \label{fig:example1}
\end{figure*}

%圖: 一張block的示意圖 一張self-attention只能apply在同一個constituent 以及group attention說明的圖
%說明motivation method大概在做什麼 為什麼會work
%Our method is built upon a intuitive assumption that the multi-head self-attention should be applied within a constituency.

Given a sentence as input, Tree Transformer induces a tree structure. %It automatically induces a tree structure from the input word sequences. 
A 3-layer Tree Transformer is illustrated in Figure~\ref{fig:example1}(A).
The building blocks of Tree Transformer is shown in Figure~\ref{fig:example1}(B), which is the same as those used in bidirectional Transformer encoder, except the proposed Constituent Attention module.
The blocks in Figure~\ref{fig:example1}(A) are constituents induced from  the input sentence.  
The red arrows indicate the self-attention. 
The words in different constituents are constrained to not attend to each other. %Tree Transformer constrains that the words only attend to the words within the same constituent.
In the $0$-th layer, some neighboring words are merged into constituents; for example, given the sentence ``\textit{the cute dog is wagging its tail}'', the tree Transformer automatically determines that ``\textit{cute}'' and ``\textit{dog}'' form a constituent, while ``\textit{its}'' and ``\textit{tail}'' also form one.
The two neighboring constituents may merge together in the next layer, so the sizes of constituents gradually grow from layer to layer.
In the top layer, the layer $2$, all words are grouped into the same constituent.
Because all words are into the same constituent, the attention heads freely attend to any other words, in this layer, Tree Transformer behaves the same as the typical Transformer encoder. %which is same as original Transformer. 
Tree Transformer can be trained in an end-to-end fashion by using ``masked LM'',  which is one of the unsupervised training task used for BERT training. 

%This architecture encourages the model learns local features such as phrase representations at lower layers, and learns global features at higher layers.
%\begin{enumerate}
%\item Add an additional module called "Constituent Attention" to original Transformer at each layer to compute the probability that the words belong to the same constituent. See Section~\ref{}.
%\item After obtaining the constituent probability at current layer, use this as prior to self-attention.
%\end{enumerate}

Whether two words belonging to the same constituent is determined by ``\emph{Constituent Prior}'' that guides the self-attention.
Constituent Prior is detailed in Section~\ref{sec:constituent_prior}, which is computed by the proposed Constituent Attention module in Section~\ref{sec:constituent_attn}.
%Because Tree Transformer is learned in an unsupervised way, with the tree structures induced by Tree Transformer, unsupervised parsing is achieved.
By using BERT masked language model as training, latent tree structures emerge from Constituent Prior and unsupervised parsing is thereby achieved.
The method for extracting the constituency parse trees from Tree Transformer is described in Section~\ref{sec:parsing}.
%The probabilities that the words belong to the same constituent called "Constituent Priors" are computed by  proposed "Constituent Attention" module, as illustrated in Figure~\ref{fig:block}.
%As illustrated in Figure~\ref{fig:block}, the Constituent Attention module takes the context from previous layer as input and output the "Constituent Prior".

%\subsection{Unsupervised Training} %Lee: 等一下再想想應該放在哪裡 -> 還是拿回原來的地方
%Lee: 這幾句話後面實驗好像講過了?
%Standard conditional language models predict the next word conditioned on current context and can only be trained from left to right.
%However, Transformer encoder is bidirectional, which allows model to peek its target words.
%BERT bypasses this problem using "Masked LM".
%It randomly masks the words in the sentence and predicts the masked words.
%Our Masked LM setup is identical to BERT that each word has 15\% to be chosen at random. 
%The 80\% of the chosen words are replaced by "[MASK]" token, 10\% of the chosen words are replaced by randomly sampled words and 10\% of chosen words remain unchanged.

%\begin{figure}[t]
%  \centering
%    \includegraphics[width=\linewidth]{tree_Transformer_block}
%      \caption{The model architecture of Tree Transformer.}
%  \label{fig:block}
%\end{figure}

%\section{Self-Attention within Constituent}
\section{Constituent Prior} \label{sec:constituent_prior}

%To allow self-attention utilize constituent structures, \cite{phrase-attention} applied the multi-head self-attention to the words within the same constituents. Lee: shall be in the related work.
%The output word representations of multi-head self-attention are regarded as constituent representations.

%Lee: 這段有點難講 ..... 原來 Transformer 算出來的東西要怎麼稱呼? self-attention weight? 那麼被 C weighted 過後又要怎麼稱呼 
%Reply: 原來的Transformer paper 只有說 "We compute the output matrix as: ... 而已，weighted後的不如稱他為weighted self-attention，然後不用給他特別的代號"

In each layer of Transformer, there are a query matrix $Q$ consisting of query vectors with dimension $d_{k}$ and a key matrix $K$ consisting of key vectors with dimension $d_{k}$.
The attention probability matrix is denoted as $E$, which is an $N$ by $N$ matrix, where $N$ is the number of words in an input sentence.
$E_{i,j}$ is the probability that the position $i$ attends to the position $j$. 
The Scaled Dot-Product Attention computes the $E$ as:
\begin{equation} \label{eq:4_pre}
%\begin{split}
E = \text{softmax}(\frac{QK^{T}}{d}),
%\end{split}
\end{equation}
where the dot-product is scaled by $1/d$.
In Transformer, the scaling factor $d$ is set to be $\sqrt{d_k}$.

%In each layer of Transformer, there are a query vector $q_i$ and a key vector $k_i$ with dimension $d_{k}$ for each position $i$.
%The attention probability $E_{i,j}$ that position $i$ attend to $j$ is computed by scaled dot product,
%\begin{equation} \label{eq:4_pre}  %Lee: 這樣講會不會太多了．．．
%Reply: 這裡有點說錯了 E 應該要是softmax後的機率才對
%\begin{split}
%E_{i,j} = softmax(\frac{q_{i}\cdot k_{j}}{d}),
%\end{split}
%\end{equation}
%which $\odot$ is the element-wise multiplication.
%Equation (\ref{eq:4_pre}) is the dot product of query vector $q_i$ and key vector $k_j$ scaled by $\frac{1}{d}$.
%In Transformer, the scaling factor $d$ is set to be $\sqrt{d_{k}}$.

In Tree Transformer, the $E$ is not only determined by the query matrix $Q$ and key matrix $K$, but also guided by Constituent Prior $C$ generating from Constituent Attention module.
Same as $E$, the constituent prior $C$ is also a $N$ by $N$ matrix, where $C_{i,j}$ is the probability that word $w_{i}$ and word $w_{j}$ belong to the same constituency.
This matrix is symmetric that $C_{i,j}$ is same as $C_{j,i}$.
Each layer has its own Constituent Prior $C$.
An example of Constituent Prior $C$ is illustrated in Figure~\ref{fig:example1}~(C), which indicates that in layer $1$, ``the cute dog'' and ``is wagging its tail'' are two constituents. 

%In tree Transformer, the constituent prior $C$ makes the words in different  constituents not attend to each other by multiplying self-attention weights in BERT with constituent priors. %we softly implement this constraint by multiplying attention probabilities by constituent priors.
To make each position not attend to the position in different constituents, Tree Transformer constrains  the attention probability matrix $E$ by constituent prior $C$ as below, 
\begin{equation} \label{eq:1}
E = C \odot \text{softmax}(\frac{QK^{T}}{d}),
\end{equation}
where $\odot$ is the element-wise multiplication.
Therefore, if $C_{i,j}$ has small value, it indicates that the positions $i$ and $j$ belong to different constituents, where the attention weight $E_{i,j}$ would be small. 
%Here we use multi-head attention as Transformer with the number of heads set to be $h$,
As Transformer uses multi-head attention with $h$ different heads, there are $h$ different query matrices $Q$ and key matrices $K$ at each position, but here in the same layer, all attention heads in multi-head attention share the same $C$.
The multi-head attention module produces the output of dimension $d_{model}=h \times d_k$. 

%The constituent prior $C$ is obtained by the consitent attention moduel in the next section.
%Following the spirit of "Attention is all you need", We obtain $C$  simply using the self-attention of two adjacent words, and thus we call it "Constituent Attention".  %先暫時放這裡

%Lee: 這個式子是沒有幫助的，因為 Q, K, V 等等在前面根本就沒有定義
%Reply: 當初這麼寫是因為原來Transformer這麼寫的，如果熟悉attention或Transformer應該馬上就能理解他的意思

%\begin{equation}  \label{eq:1}
%\begin{split}
%Attention(Q, K, V, C) = C \odot softmax(\frac{QK^{T}}{\sqrt{d_{k}}})V
%\end{split}
%\end{equation}
%, where $Q$, $K$, $V$ are query, key and value matrix respectively, $\odot$ is element-wise matrix multiplication and $d_{k}$ is the hidden dimension.

\section{Constituent Attention}\label{sec:constituent_attn} %which tiitle is better?
%\section{Constituent Priors}

%The constituent prior matrix $C$ is end-to-end learned in the tree Transformer. 
The proposed Constituent Attention module is to generate the constituent prior $C$.
Instead of directly generating $C$, we decompose the problem into estimating the breakpoints between constituents, or the probability that two adjacent words belong to the same constituent. %linking to each other.
In each layer, the Constituent Attention module generates a sequence $a=\{a_1, ..., a_i, ..., a_N\}$, where $a_{i}$ is the probability that the word $w_{i}$ and its neighbor word $w_{i+1}$ are in the same constituent.
The small value of $a_i$ implies that there is a breakpoint between $w_i$ and $w_{i+1}$, so the constituent prior $C$ is obtained from the sequence $a$ as follows.
%As directly computing constituent prior matrix $C$ of two distant words is difficult, we first decompose the problem into computing the probability of two adjacent words linking to each other.
$C_{i,j}$ is the multiplication of all $a_{i \leqslant k<j}$ between word $w_{i}$ and word $w_{j}$:
\begin{equation}  \label{eq:2}
C_{i,j} = \prod_{k=i}^{j-1} a_{k}.
\end{equation}
%This process is similar to stick-break process using in PRPN \citep{PRPN}. %Lee: 為什麼不講這句話? %因為還是有些許的不一樣 所以就沒寫
In (\ref{eq:2}), we choose to use multiplication instead of summation, because if one of $a_{i \leqslant k<j}$ between two words $w_{i}$ and $w_{j}$ is small, the value of $C_{i,j}$ with multiplication also becomes small.
In implementation, to avoid probability vanishing, we use log-sum instead of directly multiplying all $a$: 
\begin{equation}
\label{eq:3}
C_{i,j} = e^{\sum_{k=i}^{j-1} \log(a_{k})}.
\end{equation}

The sequence $a$ is obtained based on the following two mechanisms: \textit{Neighboring Attention} and \textit{Hierarchical Constraint}. 

%Neighbor Attention? It does not look like attention. Content information?
\subsection{Neighboring Attention} 
\label{sec:neighbor}

\begin{figure}[t!]
  \centering
    \includegraphics[width=\linewidth]{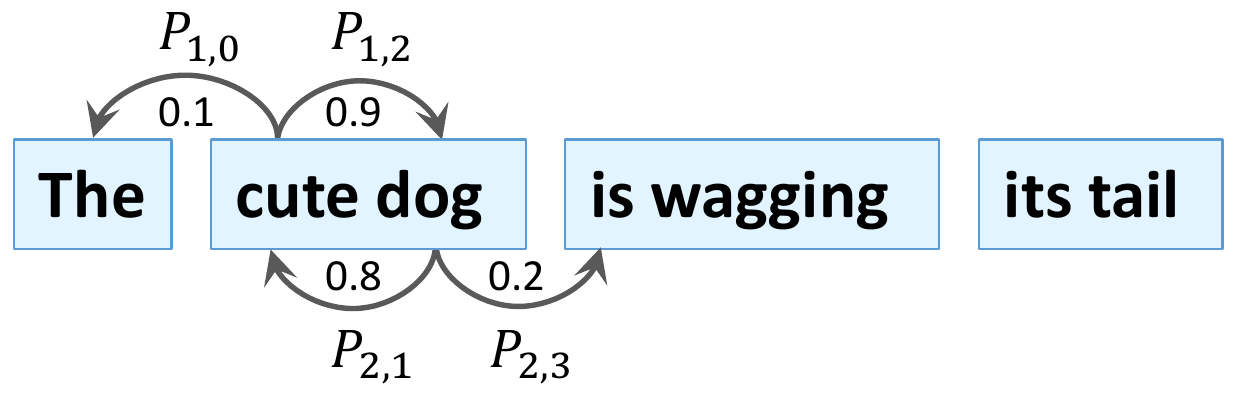}
      \caption{The example illustration about how neighboring attention works.} 
  \label{fig:example2}
\end{figure}

We compute the score $s_{i,i+1}$ indicating that $w_{i}$ links to $w_{i+1}$ by scaled dot-product attention:
\begin{equation}  \label{eq:4}
\begin{split}
s_{i,i+1} = \frac{q_{i}\cdot k_{i+1}}{d},
\end{split}
\end{equation}
where $q_{i}$ is a link query vector of $w_{i}$ with $d_{model}$ dimensions, and $k_{i+1}$ is a link key vector of $w_{i+1}$ with $d_{model}$ dimensions.
We use $q_{i}\cdot k_{i+1}$ to represent the tendency that $w_{i}$ and $w_{i+1}$ belong to the same constituent. 
Here, we set the scaling factor $d$ to be $\frac{d_{model}}{2}$.
The query and key vectors in (\ref{eq:4}) are different from (\ref{eq:4_pre}).
%$q$ and $k$ in (\label{eq:4}) are different from $\hat{q}$ and $\hat{k}$ in (\label{eq:4_pre}).
They are computed by the same network architecture, but with different sets of network parameters.

For each word, we constrain it to either link to its \emph{right neighbor} or \emph{left neighbor} as illustrated in Figure~\ref{fig:example2}.
This constraint is implemented by applying a softmax function to two attention links of $w_{i}$:
\begin{equation}  \label{eq:5}
\begin{split}
p_{i, i+1},p_{i, i-1} = \text{softmax}(s_{i,i+1},s_{i, i-1}),
\end{split}
\end{equation}
where $p_{i, i+1}$ is the probability that $w_{i}$ attends to $w_{i+1}$, and $(p_{i, i+1} + p_{i, i-1}) = 1$.
%Lee: 這裡我改一下
We find that without the constraint of the softmax operation in (\ref{eq:5}) the model prefers to link all words together and assign all words to the same constituency.
That is, giving both $s_{i,i+1}$ and $s_{i, i-1}$ large values, so the attention head freely attends to any position without restriction of constituent prior, which is the same as the original Transformer.
Therefore, the softmax function is to constraint the attention to be sparse. %to prevent this, we use softmax to ... %we have to constrain the $a$ to be sparse.

As $p_{i, i+1}$ and $p_{i+1, i}$ may have different values, we average its two attention links:
\begin{equation}  \label{eq:6}
\begin{split}
\hat{a}_{i} = \sqrt{p_{i, i+1} \times p_{i+1,i}}.
\end{split}
\end{equation}
The $\hat{a}_{i}$ links two adjacent words only if two words attend to each other.
$\hat{a}_{i}$ is used in the next subsection to obtain $a_i$.

\subsection{Hierarchical Constraint}\label{sec:update}
%\subsection{Update Constituent Priors} 
As mentioned in Section~\ref{sec:tree_Transformer}, constituents in the lower layer merge into larger one in the higher layer.  %the size of constituent grows from lower layer to higher layer.
That is, once two words belong to the same constituent in the lower layer, they would still belong to the same constituent in the higher layer.
To apply the hierarchical constraint to the tree Transformer, we restrict $a^l_k$ to be always larger than $a^{l-1}_k$ for the layer $l$ and word index $k$. %We implement this restriction by setting $C$ and $a$ strictly increasing from layer to layer.
Hence, at the layer $l$, the link probability $a^{l}_k$ is set as:
\begin{equation}  \label{eq:7}
a^{l}_k = a^{l-1}_k + (1-a^{l-1}_k) \hat{a}^{l}_k,
\end{equation}
where $a^{l-1}_k$ is the link probability from the previous layer $l-1$, and $\hat{a}^{l}_k$ is obtained from Neighboring Attention (Section~\ref{sec:neighbor}) of the current layer $l$.
Finally, at the layer $l$, we apply (\ref{eq:3}) for computing $C^{l}$ from $a^{l}$.
Initially, different words are regarded as different constituents, and thus we initialize $a^{-1}_k$ as zero.

\section{Unsupervised Parsing from Tree Transformer}\label{sec:parsing} %Constituent Attention}

%single layer
After training, the neighbor link probability $a$ can be used for unsupervised parsing.
The small value of $a$ suggests this link be the breakpoint of two constituents. 
%By recursively splitting the sentence into two constituents with minimum $a$, a parse tree can be obtained, which is called top-down greedy parsing~\citep{PRPN}.
By top-down greedy parsing~\citep{PRPN}, which recursively splits the sentence into two constituents with minimum $a$, a parse tree can be formed.

%multiple layers
%However, we have a set of $a^{l}$ at each layer $l$.
However, because each layer has a set of $a^{l}$, we have to decide to use which layer for parsing.
Instead of using $a$ from a specific layer for parsing \citep{on-lstm}, we propose a new parsing algorithm, which utilizes $a$ from all layers for unsupervised parsing.
As mentioned in Section~\ref{sec:update}, the values of $a$ are strictly increasing, which indicates that $a$ directly learns the hierarchical structures from layer to layer.
Algorithm~\ref{alg:1} details how we utilize hierarchical information of $a$ for unsupervised parsing.

The unsupervised parsing starts from the top layer, and recursively moves down to the last layer after finding a breakpoint until reaching the bottom layer $m$.
The bottom layer $m$ is a hyperparameter needed to be tuned, and is usually set to $2$ or $3$.
We discard $a$ from layers below $m$, because we find the lowest few layers do not learn good representations~\citep{BERT-ana} and thus the parsing results are poor~\citep{on-lstm}.
All values of $a$ on top few layers are very close to $1$, suggesting that those are not good breakpoints.
Therefore, we set a threshold for deciding a breakpoint, where a minimum $a$ will be viewed as a valid breakpoint only if its value is below the threshold.
As we find that our model is not very sensitive to the threshold value, we set it to be 0.8 for all experiments.

\begin{algorithm}[t!]
\caption{Unsupervised Parsing with Multiple Layers}\label{alg:1}
\begin{algorithmic}[1]
\State $a\gets$ link probabilities
\State $m \gets minimum~layer~id$\Comment{Discard the $a$ from layers below minimum layer}
\State $thres\gets 0.8$\Comment{Threshold of breakpoint}
\Procedure{BuildTree}{$l,s,e$}\Comment{$l$: layer index, $s$: start index, $e$: end index}
\If{$e - s < 2$}\Comment{The constituent cannot be split}
\State \textbf{return} $(s,e)$
\EndIf
\State $span\gets a^{l}_{s \leq i<e}$
\State $b\gets \textbf{argmin}(span)$\Comment{Get breakpoint}
\State $last\gets \textbf{max}(l-1, m)$\Comment{Get index of last layer}
\If{$a_{b}^l > thres$} %Lee: a_b -> a^l_b???%沒錯
  \If{$l = m$}
  \State \textbf{return} $(s,e)$
  \EndIf
\State \textbf{return} $\textbf{BuildTree}(last,s,e)$
\EndIf
\State $tree1\gets \textbf{BuildTree}(last,s,b)$
\State $tree2\gets \textbf{BuildTree}(last,b+1,e)$
\State \textbf{return} $(tree1,tree2)$\Comment{Return tree}
\EndProcedure
\end{algorithmic}
\end{algorithm}

\section{Experiments}

In order to evaluate the performance of our proposed model, we conduct the experiments detailed below.

\subsection{Model Architecture}
%For all experiments, we use the same model architecture.
Our model is built upon a bidirectional Transformer encoder.
The implementation of our Transformer encoder is identical to the original Transformer encoder. %except that we use a gelu activation~\citep{gelu} rather than relu, following BERT. %Lee: which one is the original encoder???
%其實用gelu結果跟relu幾乎差不多，但gelu會穩定一些些，然而我最好的結果是用relu跑的所以我就刪掉這段了
For all experiments, we set the hidden size $d_{model}$ of Constituent Attention and Transformer as 512, the number of self-attention heads $h$ as 8, the feed-forward size as 2048 and the dropout rate as 0.1.
We analyze and discuss the sensitivity of the number of layers, denoted as $L$, in the following experiments.

\subsection{Grammar Induction}

In this section, we evaluate the performance of our model on unsupervised constituency parsing.
Our model is trained on WSJ training set and WSJ-all (i.e. including testing and validation sets) by using BERT Masked LM~\citep{BERT} as unsupervised training task.
We use WordPiece~\citep{wordpiece} tokenizer from BERT to tokenize words with a 16k token vocabulary. 
Our best result is optimized by \texttt{adam} with a learning rate of 0.0001, $\beta_{1}=0.9$ and $\beta_{2}=0.98$.
%We clip the value of gradient norm to 1.5.
Following the evaluation settings of prior work \citep{htut-etal-2018-grammar-induction,on-lstm}\footnote{\url{https://github.com/yikangshen/Ordered-Neurons}}, we evaluate F1 scores of our model on WSJ-test and WSJ-10 of Penn Treebank (PTB)~\citep{Marcus:1993}. The WSJ-10 has 7422 sentences from whole PTB with sentence length restricted to 10 after punctuation removal, while WSJ-test has 2416 sentences from the PTB testing set with unrestricted sentence length.
%In Table~\ref{table:wsj-test}, we compared Tree Transformer(abbreviated as Tree-T) with recent works.

\begin{table}[t!]
\centering
\begin{tabular}{lccc}
\hline
\bf Model & \bf Data & $\textbf{F1}_{median}$ & \bf $\textbf{F1}_{max}$ \\ \hline
PRPN & WSJ-train & 35.0 & 42.8\\
On-lstm & WSJ-train & 47.7 & 49.4\\
C-PCFG & WSJ-train & \bf 55.2 & \bf 60.1\\
Tree-T,L=12 & WSJ-train & 48.4 & 50.2\\
Tree-T,L=10 & WSJ-train & 49.5 & 51.1\\
Tree-T,L=8  & WSJ-train & 48.3 & 49.6\\
Tree-T,L=6  & WSJ-train & 47.4 & 48.8\\
\hline
DIORA & NLI & \bf 55.7 & \bf 56.2\\
URNNG & Billion & - & 52.4\\
Tree-T,L=10  & WSJ-all & 50.5 & 52.0\\ \hline
Random & - & 21.6 & 21.8\\
LB & - & 9.0 & 9.0\\
RB & - & 39.8 & 39.8\\ \hline
\end{tabular}
\caption{\textbf{The F1 scores on WSJ-test.} Tree Transformer is abbreviated as Tree-T, and L is the number of layers(blocks). DIORA is trained on multi-NLI dataset~\citep{williams-etal-2018-broad}. URNNG is trained on the subset of one billion words~\citep{billion} with 1M training data. LB and RB are the left and right-brancing baselines.}
\label{table:wsj-test}
\end{table}

The results on WSJ-test are in Table~\ref{table:wsj-test}.
%Tree Transformer performs better than On-lstm and PRPN, in which the evaluation settings are same as our model.
%Compared to DIORA or URNNG, we use a relative small training data to train our model, showing the better capability of maintaining the tree structure in our model.
We mainly compare our model to PRPN~\citep{PRPN}, On-lstm~\citep{on-lstm} and Compound PCFG(C-PCFG)~\citep{CPCFG}, in which the evaluation settings and the training data are identical to our model.
DIORA~\citep{diora} and URNNG~\citep{urnng} use a relative larger training data and the evaluation settings are slightly different from our model.
%In random baseline, we randomly initialize the parameters of our model and evaluate its results. %Lee: this obtains random tree? %是的做出來也跟on-lstm report的random tree差不多 但是這應該不用特別敘述
Our model performs much better than trivial trees (i.e. right and left-branching trees) and random trees, which suggests that our proposed model successfully learns meaningful trees.
We also find that increasing the layer number results in better performance, because it allows the Tree Transformer to model deeper trees.
However, the performance stops growing when the depth is above 10.
The words in the layers above the certain layer are all grouped into the same constituent, and therefore increasing the layer number will no longer help model discover useful tree structures.
In Table~\ref{table:wsj-10}, we report the results on WSJ-10.
Some of the baselines including CCM~\citep{klein-manning-2002-generative}, DMV+CCM~\citep{Klein:2005:NLG:1746577.1746604} and UML-DOP~\citep{Bod:2006:AAU:1220175.1220284} are not directly comparable to our model, because they are trained using POS tags our model does not consider.

\begin{table}[t!]
\centering
\begin{tabular}{lccc}
\hline
\bf Model & \bf Data & $\textbf{F1}_{median}$ & \bf $\textbf{F1}_{max}$ \\ \hline
PRPN & WSJ-train & 70.5 &  71.3\\
On-lstm & WSJ-train & 65.1 & 66.8\\
C-PCFG & WSJ-train & 70.5 & -\\
Tree-T,L=10 & WSJ-train & 66.2 & 67.9\\
\hline
DIORA & NLI & 67.7 & 68.5\\
Tree-T,L=10  & WSJ-all & 66.2 & 68.0\\
 \hline
CCM & WSJ-10 & - & 71.9\\
DMV+CCM & WSJ-10 & - & 77.6\\
UML-DOP & WSJ-10 & - & \bf 82.9\\ \hline
Random & - & 31.9 & 32.6\\
LB & - & 19.6 & 19.6\\
RB & - & 56.6 & 56.6\\ \hline
\end{tabular}
\caption{\textbf{The F1 scores on WSJ-10.} Tree Transformer is abbreviated as Tree-T, and L is the number of layers (blocks).}
\label{table:wsj-10}
\end{table}

\begin{table}[t!]
\centering
\begin{tabular}{lccc}
\hline
\bf Label & \bf Tree-T & \bf URNNG & \bf PRPN \\
\hline
NP & \bf 67.6 & 39.5 & 63.9 \\
VP & 38.5 & \bf 76.6 & 27.3 \\
PP & 52.3 & \bf 55.8 & 55.1 \\
ADJP & 24.7 & 33.9 & \bf 42.5 \\
SBAR & 36.4 & \bf 74.8 & 28.9 \\
ADVP & \bf 55.1 & 50.4 & 45.1 \\
\hline
\end{tabular}
\caption{\textbf{Recall of constituents by labels.} The results of URNNG and PRPN are taken from~\citet{urnng}. }
\label{table:constituent}
\end{table}
%The unsupervised constituency parsing is the task of learning latent tree structures without human expert annotated data.

\begin{figure}[t!]
  \centering
    \includegraphics[width=\linewidth]{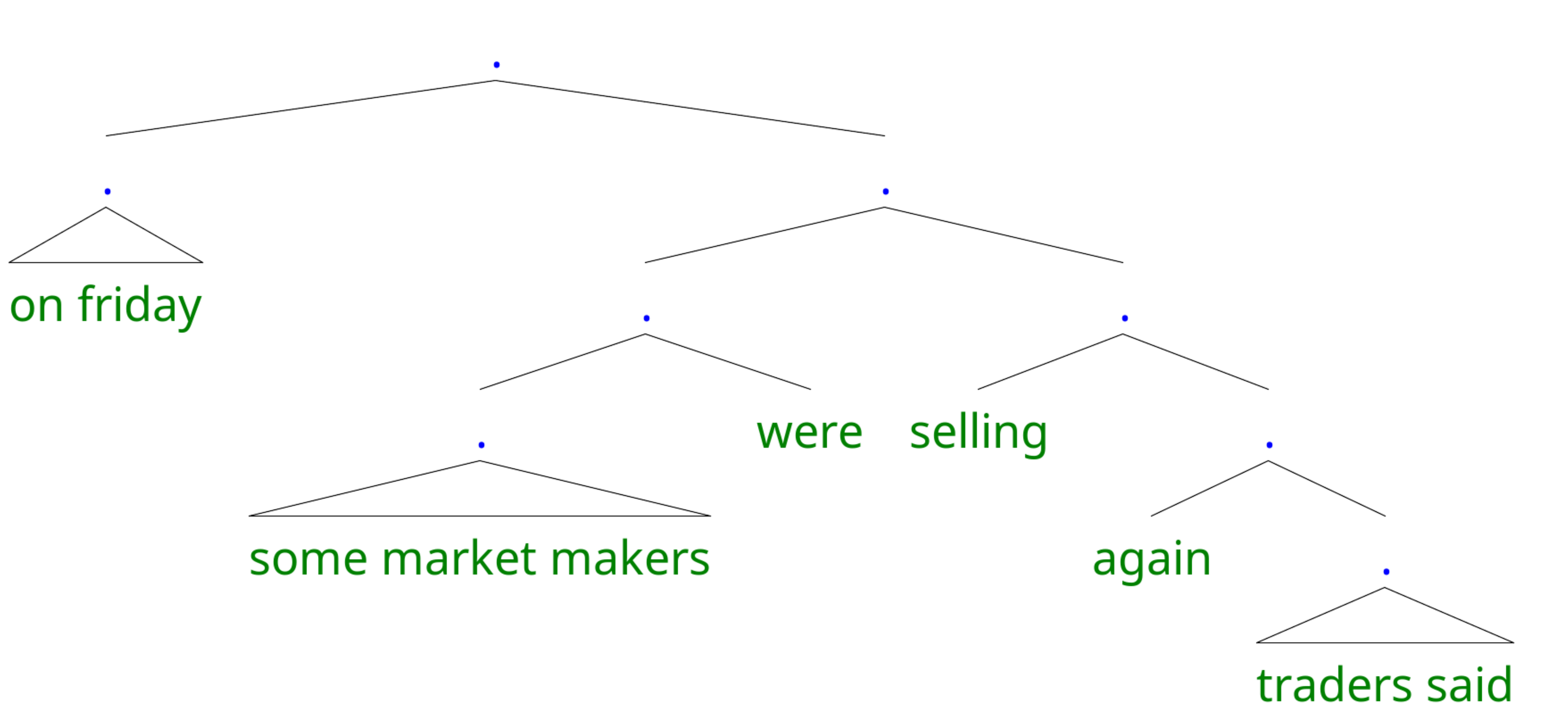}
      \caption{A parse tree induced by Tree Transformer. As shown in the figure, because we set a threshold in Algorithm~\ref{alg:1}, the leaf nodes are not strictly binary.}
  \label{fig:tree1}
\end{figure}

In addition, we further investigate what kinds of trees are induced by our model.
Following URNNG, we evaluate the performance of constituents by its label in Table~\ref{table:constituent}.
The trees induced by different methods are quite different.
Our model is inclined to discover noun phrases (NP) and adverb phrases (ADVP), but not easy to discover verb phrases (VP) or adjective phrases (ADJP).
We show an induced parse tree in Figure~\ref{fig:tree1} and more induced parse trees can be found in Appendix.
%討論大概學到了什麼parse tree
%討論layer的數目

\begin{figure*}[t!]
\centering  
\subfigure[F1 in terms of different minimum layers.]{\includegraphics[width=0.49\linewidth]{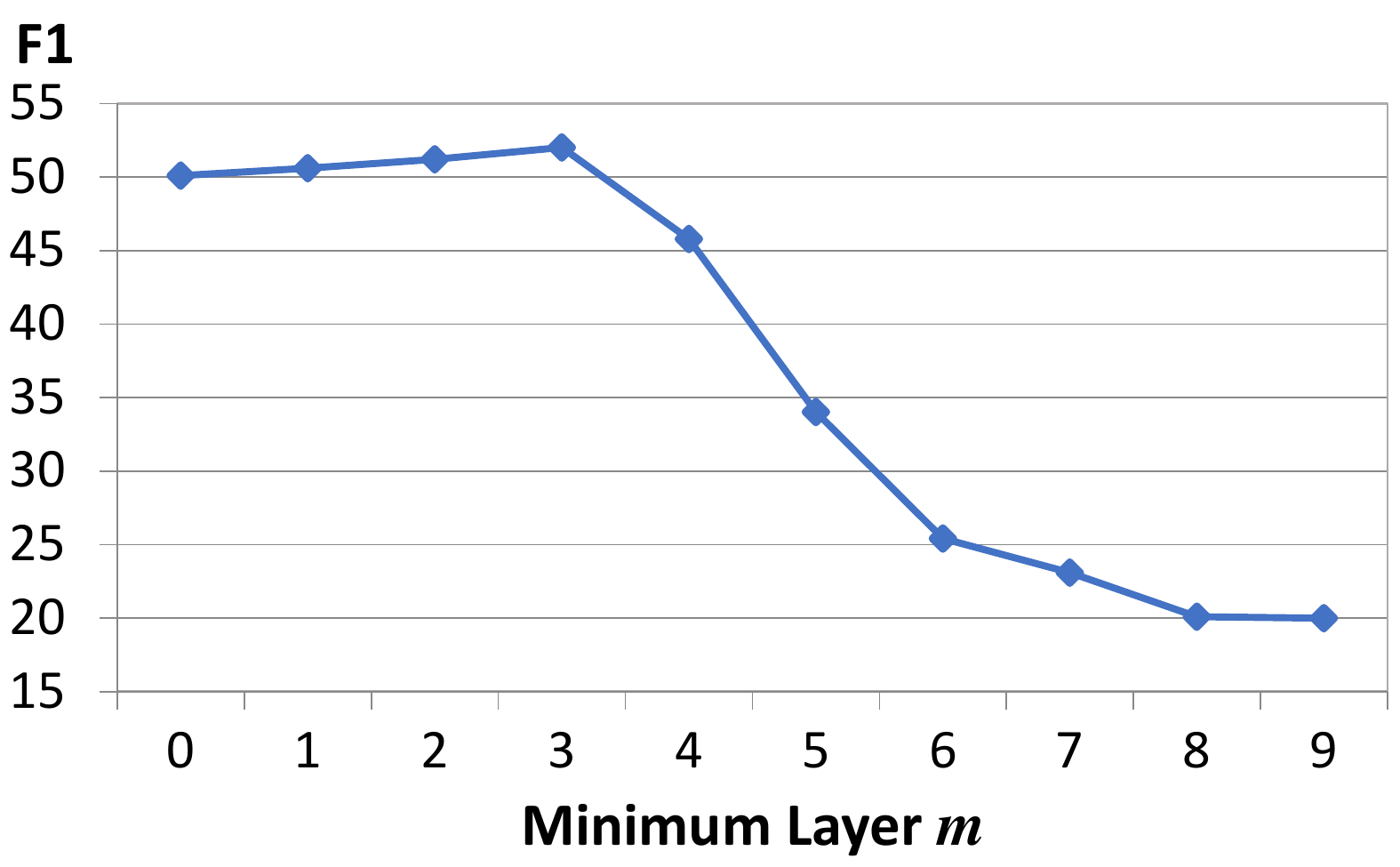}}
\subfigure[F1 of parsing via a specific layer.]{\includegraphics[width=0.49\linewidth]{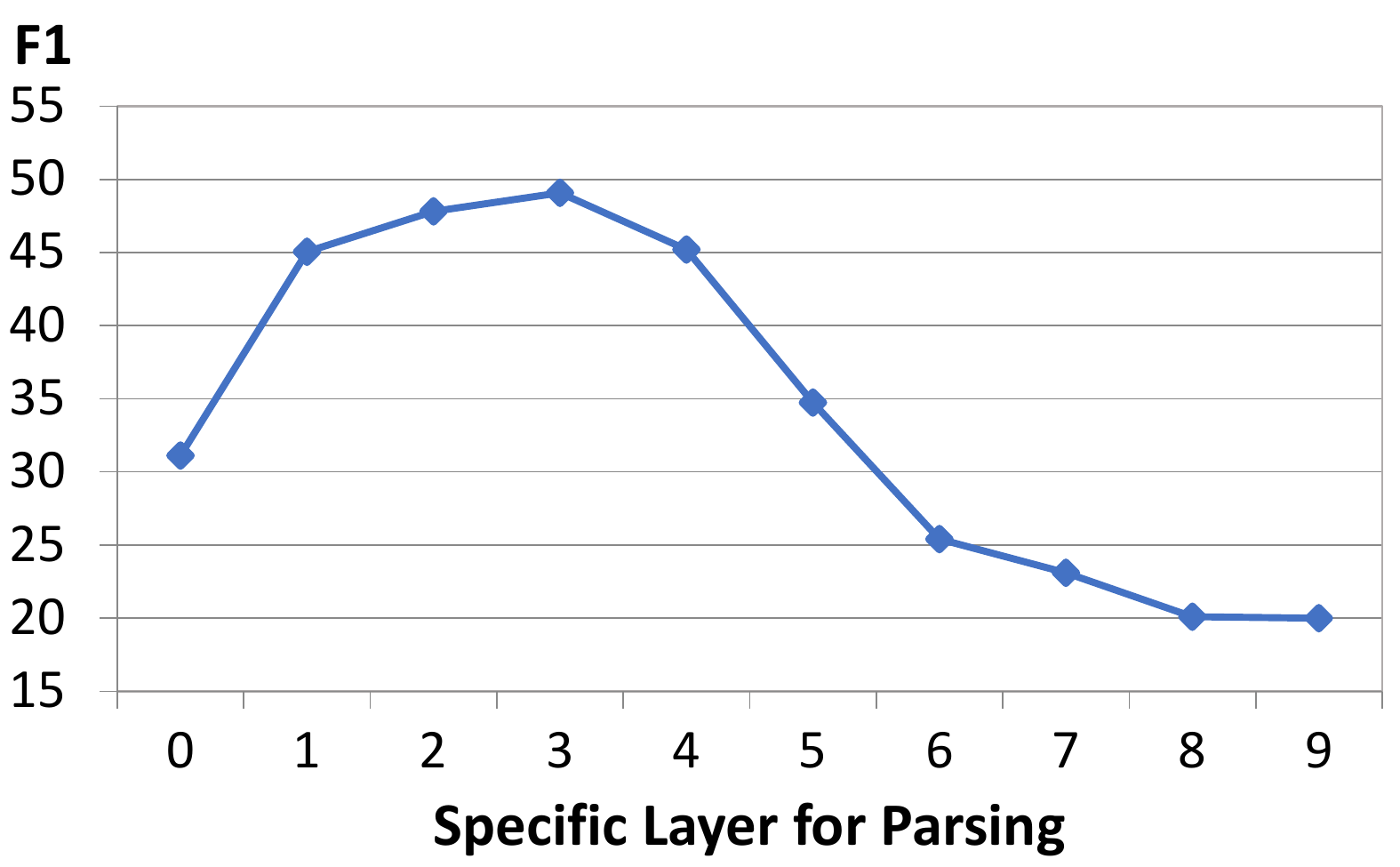}}
\caption{Performance of unsupervised parsing.} 
  \label{fig:F1}
\end{figure*}

\iffalse
\begin{figure}[t!]
  \centering
    \includegraphics[width=\linewidth]{F1-m}
      %\setlength{\abovecaptionskip}{-10pt}
      %\setlength{\textfloatsep}{-10pt}
      \caption{Performance of different bottom layers.} 
  \label{fig:different_layer}
\end{figure}

\begin{figure}[t!]
  \centering
    \includegraphics[width=\linewidth]{F1-layer}
      %\setlength{\abovecaptionskip}{-10pt}
      %\setlength{\belowcaptionskip}{-10pt}
      \caption{Performance of parsing via a specific layer.} 
  \label{fig:specific_layer}
\end{figure}
\fi

\subsection{Analysis of Induced Structures}
In this section, we study whether Tree Transformer learns hierarchical structures from layers to layers.
First, we analyze the influence of the hyperparameter minimum layer $m$ in Algorithm.~\ref{alg:1} given the model trained on WSJ-all in Table~\ref{table:wsj-test}.
As illustrated in Figure~\ref{fig:F1}(a), setting $m$ to be 3 yields the best performance.
Prior work discovered that the representations from the lower layers of Transformer are not informative~\cite{BERT-ana}.
Therefore, using syntactic structures from lower layers decreases the quality of parse trees.
On the other hand, most syntactic information is missing when $a$ from top few layers are close to $1$, so too large $m$ also decreases the performance. %Lee:這句看不懂 %已經修改了，也可以考慮拿掉?

To further analyze which layer contains richer information of syntactic structures, we evaluate the performance on obtaining parse trees from a specific layer.
We use $a^{l}$ from the layer $l$ for parsing with the top-down greedy parsing algorithm~\citep{PRPN}.
As shown in Figure~\ref{fig:F1}(b), using $a^{3}$ from the layer 3 for parsing yields the best F1 score, which is $49.07$.
The result is consistent to the best value of $m$.
However, compared to our best result ($52.0$) obtained by Algorithm~\ref{alg:1}, the F1-score decreases by $3$ ($52.0 \rightarrow 49.07$).
This demonstrates the effectiveness of Tree Transformer in terms of learning hierarchical structures.
The higher layers indeed capture the higher-level syntactic structures such as clause patterns.

\subsection{Interpretable Self-Attention}
%\subsection{Visualization of self-attention heads}

%In this section we discuss whether the attention heads in Tree Transformer learns hierarchical structures by visualizing the attention scores: $C \odot softmax(\frac{QK^{T}}{\sqrt{d_{k}}})$ in eq~\ref{eq:1}.

\begin{figure*}[t!]
\centering  
\subfigure[The layer 0.]{\includegraphics[width=0.495\linewidth]{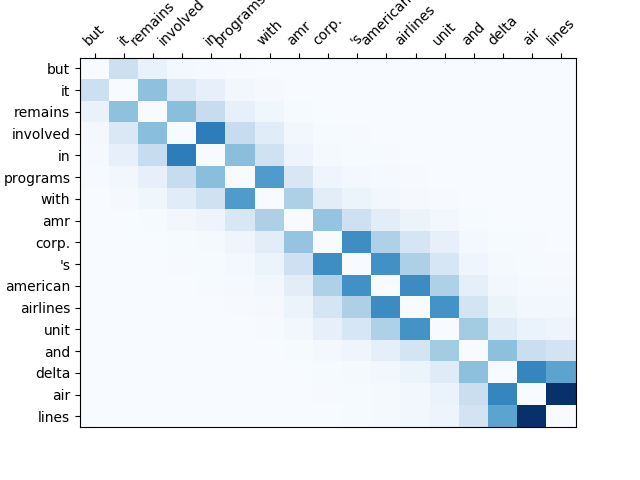}}
\subfigure[The layer 3.]{\includegraphics[width=0.495\linewidth]{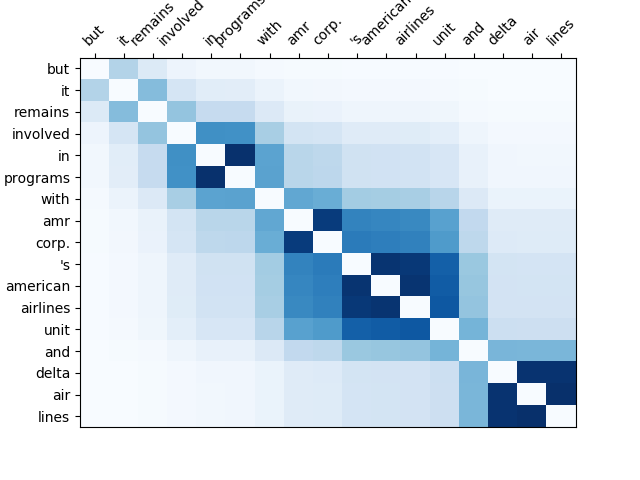}}
\subfigure[The layer 6.]{\includegraphics[width=0.495\linewidth]{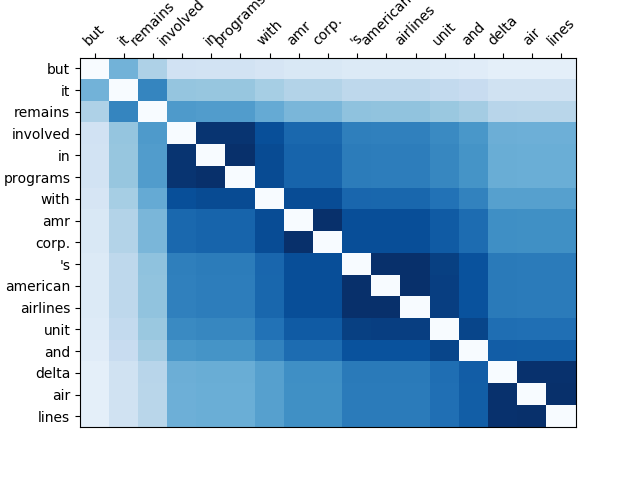}}
\subfigure[The layer 9.]{\includegraphics[width=0.495\linewidth]{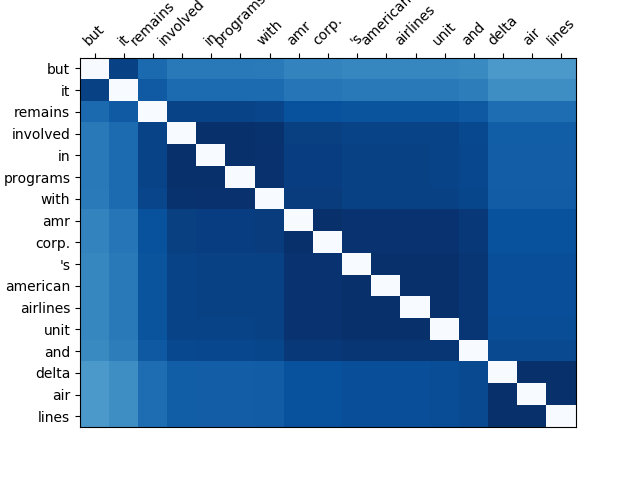}}
\caption{The constituent prior heat maps.} 
  \label{fig:g_attn}
\end{figure*}

This section discusses whether the attention heads in Tree Transformer learn hierarchical structures.
Considering that the most straightforward way of interpreting what attention heads learn is to visualize the attention scores, we plot the heat maps of Constituent Attention prior $C$ from each layer in Figure~\ref{fig:g_attn}. %we plot the heat map of attention scores $C \odot softmax(\frac{QK^{T}}{\sqrt{d_{k}}})$ in eq~\ref{eq:1}.
%Because we want to demonstrate Tree Transformer learns layer-wise tree structures, we average the scores of the attention heads within the same layer.

In the heat map of constituent prior from first layer  (Figure~\ref{fig:g_attn}(a)), as the size of constituent is small, the words only attend to its adjacent words.
We can observe that the model captures some sub-phrase structures, such as the noun phrase ``\emph{delta air line}'' or ``\emph{american airlines unit}''.
In Figure~\ref{fig:g_attn}(b)-(d), the constituents attach to each other and become larger.
In the layer 6, the words from ``\emph{involved}'' to last word ``\emph{lines}'' form a high-level adjective phrase (ADJP).
In the layer 9, all words are grouped into a large constituent except the first word ``\emph{but}''.
By visualizing the heat maps of the constituent prior from each layer, we can easily know what types of syntactic structures are learned in each layer.
The parse tree of this example can be found in Figure~\ref{fig:tree1} of Appendix.
We also visualize the heat maps of self-attention from the original Transformer layers and one from the Tree Transformer layers in Appendix~\ref{app:a}.
As the self-attention heads from our model are constrained by the constituent prior, compared to the original Transformer, we can discover hierarchical structures more easily. 
%We also observe that the words within the same phrase attend to each other, such as phrasal verb "involve in" and noun phrase "delta air line".
%With the constituents attaching to each other and becoming larger, in Figure~\ref{fig:layer3}, Figure~\ref{fig:layer6} and Figure~\ref{fig:layer6}, the words attend to more distant words.

%========================================這段是否放到appendix比較合適啊，目前版面不夠
%On the other hard, we average the scores of attention heads from each layer of original Transformer and visualize them.
%In the attention heat maps of original Transformer, we also observe it learns hierarchical structures in the first layer(Figure~\ref{fig:transformer_layer0}) that the words mostly attend to its adjacent words.
%However, this phenomenon is only observed in the first layer that in third layer(Figure~\ref{fig:transformer_layer3}), the attention heads attend to other words at will.
%In addition, we also provide the weighted attention scores $C \odot softmax(\frac{QK^{T}}{\sqrt{d_{k}}})$ in Appendix.\ref{app:a}. %Lee: I will change the notation
%==========================================

Those attention heat maps demonstrate that: (1) the size of constituents gradually grows from layer to layer, and (2) at each layer, the attention heads tend to attend to other words within constituents posited in that layer.
Those two evidences support the success of the proposed Tree Transformer in terms of learning tree-like structures.

%To show the attention heads gradually attend to more distant words from layer to layer in statistic, we evaluate the attention distance at each layer.
%The attention distance of a sentence is $\sum D_{i,j}S_{i,j}$, where $D_{i,j}=\left | i-j \right |$ is the distance of $w_{i}$ and $w_{j}$ and $S_{i,j}$ is the average of attention scores from all attention heads within the same layer.

%還可以分析的: Constituent Attention的visualize
%Constituent Attention的heat map

\begin{table}[t!]
\centering
\begin{tabular}{lccc}
\hline
\bf Model & \bf L & \bf Params & \bf Perplexity \\ \hline
Transformer & 8 & 40M & 48.8 \\
Transformer & 10 & 46M & 48.5 \\
Transformer & 10-B & 67M & 49.2 \\
Transformer & 12 & 52M & 48.1 \\ \hline
Tree-T & 8 & 44M & 46.1\\
Tree-T & 10 & 51M & 45.7\\
Tree-T & 12 & 58M & 45.6\\
\hline
\end{tabular}
\caption{\textbf{The perplexity of masked words.} Params is the number of parameters. We denote the number of layers as $L$. In Transformer $L=10-B$, the increased hidden size results in more parameters.}
\label{table:masked_lm}
\end{table}

\subsection{Masked Language Modeling} %Lee: 這段放到後面比較合適

%Language modeling evaluates the capability of a model to capture abstract concept and syntactic knowledge.
To investigate the capability of Tree Transformer in terms of capturing abstract concepts and syntactic knowledge, we evaluate the performance on language modeling.
As our model is a bidirectional encoder, in which the model can see its subsequent words, we cannot evaluate the language model in a left-to-right manner.
We evaluate the performance on masked language modeling by measuring the perplexity on masked words\footnote{The perplexity of masked words is $e^{\frac{-\sum \log(p)}{n_{mask}}}$, where $p$ is the probability of correct masked word to be predicted and $n_{mask}$ is the total number of masked words.}.
To perform the inference without randomness, for each sentence in the testing set, we mask all words in the sentence, but not at once.
In each masked testing data, only one word is replaced with a ``\textsf{[MASK]}'' token.
Therefore, each sentence creates the number of testing samples equal to its length.

\iffalse
\begin{figure}[t]
  \centering
    \includegraphics[width=\linewidth]{g_attn_layer_0.png}
      \setlength{\abovecaptionskip}{-10pt}
      \caption{The constituent prior heat map from layer 0.} 
  \label{fig:g_attn0}
\end{figure}

\begin{figure}[t]
  \centering
    \includegraphics[width=\linewidth]{g_attn_layer_3.png}
      \setlength{\abovecaptionskip}{-10pt}
      \caption{The constituent prior heat map from layer 3.}
  \label{fig:g_attn3}
\end{figure}

\begin{figure}[t]
  \centering
    \includegraphics[width=\linewidth]{g_attn_layer_6.png}
     \setlength{\abovecaptionskip}{-10pt}
      \caption{The constituent prior heat map from layer 6.} 
  \label{fig:g_attn6}
\end{figure}

\begin{figure}[t]
  \centering
    \includegraphics[width=\linewidth]{g_attn_layer_9.png}
      \setlength{\abovecaptionskip}{-10pt}
      \caption{The constituent prior heat map from layer 9.}
  \label{fig:g_attn9}
\end{figure}
\fi

In Table~\ref{table:masked_lm}, the models are trained on WSJ-train with BERT masked LM and evaluated on WSJ-test.
All hyperparameters except the number of layers in Tree Transformer and Transformer are set to be the same and optimized by the same optimizer.
We use \texttt{adam} as our optimizer with learning rate of 0.0001, $\beta_{1}=0.9$ and $\beta_{2}=0.999$.
%We clip the value of gradient norm to 1.5.
Our proposed Constituent Attention module increases about $10\%$ hyperparameters to the original Transformer encoder and the computational speed is 1.2 times slower.
The results with best performance on validation set are reported.
Compared to the original Transformer, Tree Transformer achieves better performance on masked language modeling.
As the performance gain is possibly due to more parameters, we adjust the number of layers or increase the number of hidden layers in Transformer $L=10-B$.
Even with fewer parameters than Transformer, Tree Transformer still performs better.

The performance gain is because the induced tree structures guide the self-attention processes language in a more straightforward and human-like manner, and thus the knowledge can be better generalized from training data to testing data.
Also, Tree Transformer acquires positional information not only from positional encoding but also from the induced tree structures, where the words attend to other words from near to distant (lower layers to higher layers)\footnote{We do not remove the positional encoding from the Transformer and we find that without positional encoding the quality of induced parse trees drops.}.
%Also, the tree structures help Transformer acquire the relative positional information that from lower layers to higher layers, the words attend to other words from near to distant.

%\subsection{}
%l=8, 57.6
%l=12=61.5
%l=10=62.4
%To evaluate whether our induced tree structures can benefit downstream tasks, we evaluate Tree Transformer on MultiNLI dataset. MultiNLI is a ????? task. When conducting experiments on this dataset, we concatenate two input sentences as input sequence and add a special tag ``[CLS]'' at the beginning of the input sequence. Then, we feed the representation of the special tag into a fully-connected neural network and get the classification result.  
%

\subsection{Limitations and Discussion} 
It is worth mentioning that we have tried to initialize our Transformer model with pre-trained BERT, and then fine-tuning on WSJ-train.
However, in this setting, even when the training loss becomes lower than the loss of training from scratch, the parsing result is still far from our best results.
This suggests that the attention heads in pre-trained BERT learn quite different structures from the tree-like structures in Tree Transformer.
%The attention heads of pre-trained BERT only~\citep{visualize-Transformer}
In addition, with a well-trained Transformer, it is not necessary for the Constituency Attention module to induce reasonable tree structures, because the training loss decreases anyway.

\section{Conclusion}
This paper proposes Tree Transformer, a first attempt of integrating tree structures into Transformer by constraining the attention heads to attend within constituents.
The tree structures are automatically induced from the raw texts by our proposed Constituent Attention module, which attaches the constituents to each other by self-attention.
The performance on unsupervised parsing demonstrates the effectiveness of our model in terms of inducing tree structures coherent to human expert annotations.
We believe that incorporating tree structures into Transformer is an important and worth exploring direction, because it allows Transformer to learn more interpretable attention heads and achieve better language modeling. 
The interpretable attention can better explain how the model processes the natural language and guide the future improvement.
%, which is highly possible to be beneficial to downstream tasks.

\section*{Acknowledgements}
We would like to thank reviewers for their insightful comments. This work was financially supported from the Young Scholar Fellowship Program by Ministry of Science and Technology (MOST) in Taiwan, under Grant 108-2636-E002-003.

\bibliography{emnlp-ijcnlp-2019}

\begin{thebibliography}{30}
\expandafter\ifx\csname natexlab\endcsname\relax\def\natexlab#1{#1}\fi

\bibitem[{Aharoni and Goldberg(2017)}]{aharoni-goldberg-2017-towards}
Roee Aharoni and Yoav Goldberg. 2017.
\newblock \href {https://www.aclweb.org/anthology/P17-2021} {Towards
  string-to-tree neural machine translation}.
\newblock In \emph{Proceedings of the 55th Annual Meeting of the Association
  for Computational Linguistics (Volume 2: Short Papers)}.

\bibitem[{Bod(2006)}]{Bod:2006:AAU:1220175.1220284}
Rens Bod. 2006.
\newblock \href {https://doi.org/10.3115/1220175.1220284} {An all-subtrees
  approach to unsupervised parsing}.
\newblock In \emph{Proceedings of the 21st International Conference on
  Computational Linguistics and the 44th Annual Meeting of the Association for
  Computational Linguistics}.

\bibitem[{Carroll and Charniak(1992)}]{Carroll92twoexperiments}
Glenn Carroll and Eugene Charniak. 1992.
\newblock Two experiments on learning probabilistic dependency grammars from
  corpora.
\newblock In \emph{WORKING NOTES OF THE WORKSHOP STATISTICALLY-BASED NLP
  TECHNIQUES}.

\bibitem[{C.Goller and A.Kuchler(1996)}]{goller}
C.Goller and A.Kuchler. 1996.
\newblock Learning task-dependent distributed representations by
  backpropagation through structure.
\newblock In \emph{Proceedings of International Conference on Neural Networks
  (ICNN'96)}.

\bibitem[{Chelba et~al.(2013)Chelba, Mikolov, Schuster, Ge, Brants, Koehn, and
  Robinson}]{billion}
Ciprian Chelba, Tomas Mikolov, Mike Schuster, Qi~Ge, Thorsten Brants, Phillipp
  Koehn, and Tony Robinson. 2013.
\newblock One billion word benchmark for measuring progress in statistical
  language modeling.
\newblock \emph{arXiv preprint arXiv:1312.3005}.

\bibitem[{Devlin et~al.(2018)Devlin, Chang, Lee, and Toutanova}]{BERT}
Jacob Devlin, Ming-Wei Chang, Kenton Lee, and Kristina Toutanova. 2018.
\newblock Bert: Pre-training of deep bidirectional transformers for language
  understanding.
\newblock \emph{arXiv preprint arXiv:1810.04805}.

\bibitem[{Dong et~al.(2019)Dong, Yang, Wang, Wei, Liu, Wang, Gao, Zhou, and
  Hon}]{BERT-generation}
Li~Dong, Nan Yang, Wenhui Wang, Furu Wei, Xiaodong Liu, Yu~Wang, Jianfeng Gao,
  Ming Zhou, and Hsiao-Wuen Hon. 2019.
\newblock Unified language model pre-training for natural language
  understanding and generation.
\newblock \emph{arXiv preprint arXiv:1905.03197}.

\bibitem[{Drozdov et~al.(2019)Drozdov, Verga, Yadav, Iyyer, and
  McCallum}]{diora}
Andrew Drozdov, Pat Verga, Mohit Yadav, Mohit Iyyer, and Andrew McCallum. 2019.
\newblock Unsupervised latent tree induction with deep inside-outside recursive
  autoencoders.
\newblock In \emph{North American Association for Computational Linguistics}.

\bibitem[{Dyer et~al.(2016)Dyer, Kuncoro, Ballesteros, and
  Smith}]{dyer-rnng:16}
Chris Dyer, Adhiguna Kuncoro, Miguel Ballesteros, and Noah~A. Smith. 2016.
\newblock Recurrent neural network grammars.
\newblock In \emph{Proc. of NAACL}.

\bibitem[{Eriguchi et~al.(2017)Eriguchi, Tsuruoka, and
  Cho}]{eriguchi-etal-2017-learning}
Akiko Eriguchi, Yoshimasa Tsuruoka, and Kyunghyun Cho. 2017.
\newblock \href {https://www.aclweb.org/anthology/P17-2012} {Learning to parse
  and translate improves neural machine translation}.
\newblock In \emph{Proceedings of the 55th Annual Meeting of the Association
  for Computational Linguistics (Volume 2: Short Papers)}.

\bibitem[{Htut et~al.(2018)Htut, Cho, and
  Bowman}]{htut-etal-2018-grammar-induction}
Phu~Mon Htut, Kyunghyun Cho, and Samuel Bowman. 2018.
\newblock \href {https://www.aclweb.org/anthology/W18-5452} {Grammar induction
  with neural language models: An unusual replication}.
\newblock In \emph{Proceedings of the 2018 {EMNLP} Workshop {B}lackbox{NLP}:
  Analyzing and Interpreting Neural Networks for {NLP}}. Association for
  Computational Linguistics.

\bibitem[{Kim et~al.(2019{\natexlab{a}})Kim, Dyer, and Rush}]{CPCFG}
Yoon Kim, Chris Dyer, and Alexander Rush. 2019{\natexlab{a}}.
\newblock \href {https://www.aclweb.org/anthology/P19-1228} {Compound
  probabilistic context-free grammars for grammar induction}.
\newblock In \emph{Proceedings of the 57th Annual Meeting of the Association
  for Computational Linguistics}, pages 2369--2385.

\bibitem[{Kim et~al.(2019{\natexlab{b}})Kim, Rush, Yu, Kuncoro, Dyer, and
  Melis}]{urnng}
Yoon Kim, Alexander~M. Rush, Lei Yu, Adhiguna Kuncoro, Chris Dyer, and Gábor
  Melis. 2019{\natexlab{b}}.
\newblock Unsupervised recurrent neural network grammars.
\newblock \emph{arXiv preprint arXiv:1904.03746}.

\bibitem[{Klein and Manning(2002)}]{klein-manning-2002-generative}
Dan Klein and Christopher~D. Manning. 2002.
\newblock \href {https://www.aclweb.org/anthology/P02-1017} {A generative
  constituent-context model for improved grammar induction}.
\newblock In \emph{Proceedings of 40th Annual Meeting of the Association for
  Computational Linguistics}.

\bibitem[{Klein and Manning(2005)}]{Klein:2005:NLG:1746577.1746604}
Dan Klein and Christopher~D. Manning. 2005.
\newblock \href {http://dx.doi.org/10.1016/j.patcog.2004.03.023} {Natural
  language grammar induction with a generative constituent-context model}.
\newblock \emph{Pattern Recogn.}

\bibitem[{Liu et~al.(2019)Liu, Gardner, Belinkov, Peters, and Smith}]{BERT-ana}
Nelson~F. Liu, Matt Gardner, Yonatan Belinkov, Matthew~E. Peters, and Noah~A.
  Smith. 2019.
\newblock Linguistic knowledge and transferability of contextual
  representations.
\newblock \emph{arXiv preprint arXiv:1903.08855}.

\bibitem[{Marcus et~al.(1993)Marcus, Marcinkiewicz, and
  Santorini}]{Marcus:1993}
Mitchell~P. Marcus, Mary~Ann Marcinkiewicz, and Beatrice Santorini. 1993.
\newblock \href {http://dl.acm.org/citation.cfm?id=972470.972475} {Building a
  large annotated corpus of english: The penn treebank}.
\newblock \emph{Comput. Linguist.}

\bibitem[{Radford et~al.(2019)Radford, Wu, Child, Luan, Amodei, and
  Sutskever}]{GPT-2}
Alec Radford, Jeff Wu, Rewon Child, David Luan, Dario Amodei, and Ilya
  Sutskever. 2019.
\newblock Language models are unsupervised multitask learners.

\bibitem[{Shen et~al.(2018{\natexlab{a}})Shen, Lin, wei Huang, and
  Courville}]{PRPN}
Yikang Shen, Zhouhan Lin, Chin wei Huang, and Aaron Courville.
  2018{\natexlab{a}}.
\newblock \href {https://openreview.net/forum?id=rkgOLb-0W} {Neural language
  modeling by jointly learning syntax and lexicon}.
\newblock In \emph{International Conference on Learning Representations}.

\bibitem[{Shen et~al.(2018{\natexlab{b}})Shen, Tan, Sordoni, and
  Courville}]{on-lstm}
Yikang Shen, Shawn Tan, Alessandro Sordoni, and Aaron Courville.
  2018{\natexlab{b}}.
\newblock Ordered neurons: Integrating tree structures into recurrent neural
  networks.
\newblock \emph{arXiv preprint arXiv:1810.09536}.

\bibitem[{Smith and Eisner(2005)}]{Smith05guidingunsupervised}
Noah~A. Smith and Jason Eisner. 2005.
\newblock Guiding unsupervised grammar induction using contrastive estimation.
\newblock In \emph{In Proc. of IJCAI Workshop on Grammatical Inference
  Applications}.

\bibitem[{Socher et~al.(2011)Socher, Lin, Ng, and
  Manning}]{Socher:2011:PNS:3104482.3104499}
Richard Socher, Cliff Chiung-Yu Lin, Andrew~Y. Ng, and Christopher~D. Manning.
  2011.
\newblock Parsing natural scenes and natural language with recursive neural
  networks.
\newblock In \emph{Proceedings of the 28th International Conference on
  International Conference on Machine Learning}.

\bibitem[{Strubell et~al.(2018)Strubell, Verga, Andor, Weiss, and
  McCallum}]{Strubell2018LinguisticallyInformedSF}
Emma Strubell, Patrick Verga, Daniel Andor, David~I Weiss, and Andrew McCallum.
  2018.
\newblock Linguistically-informed self-attention for semantic role labeling.
\newblock In \emph{EMNLP}.

\bibitem[{Tai et~al.(2015)Tai, Socher, and Manning}]{tree-lstm}
Kai~Sheng Tai, Richard Socher, and Christopher~D. Manning. 2015.
\newblock Improved semantic representations from tree-structured long
  short-term memory networks.
\newblock \emph{arXiv preprint arXiv:1503.00075}.

\bibitem[{Vig(2019)}]{visualize-Transformer}
Jesse Vig. 2019.
\newblock Visualizing attention in transformer-based language representation
  models.
\newblock \emph{arXiv preprint arXiv:1904.02679}.

\bibitem[{Williams et~al.(2018)Williams, Nangia, and
  Bowman}]{williams-etal-2018-broad}
Adina Williams, Nikita Nangia, and Samuel Bowman. 2018.
\newblock \href {https://www.aclweb.org/anthology/N18-1101} {A broad-coverage
  challenge corpus for sentence understanding through inference}.
\newblock In \emph{Proceedings of the 2018 Conference of the North {A}merican
  Chapter of the Association for Computational Linguistics: Human Language
  Technologies, Volume 1 (Long Papers)}.

\bibitem[{Wu et~al.(2018)Wu, Wang, Liu, and Ma}]{phrase-attention}
Wei Wu, Houfeng Wang, Tianyu Liu, and Shuming Ma. 2018.
\newblock \href {https://www.aclweb.org/anthology/D18-1408} {Phrase-level
  self-attention networks for universal sentence encoding}.
\newblock In \emph{Proceedings of the 2018 Conference on Empirical Methods in
  Natural Language Processing}, pages 3729--3738.

\bibitem[{Wu et~al.(2016)Wu, Schuster, Chen, Le, Norouzi, Macherey, Krikun,
  Cao, Gao, Macherey, and et~al.}]{wordpiece}
Yonghui Wu, Mike Schuster, Zhifeng Chen, Quoc~V. Le, Mohammad Norouzi, Wolfgang
  Macherey, Maxim Krikun, Yuan Cao, Qin Gao, Klaus Macherey, and et~al. 2016.
\newblock \href {http://arxiv.org/abs/1609.08144} {Google's neural machine
  translation system: Bridging the gap between human and machine translation}.
\newblock \emph{arXiv preprint arXiv:1609.08144}.

\bibitem[{Yogatama et~al.(2017)Yogatama, Blunsom, Dyer, Grefenstette, and
  Ling}]{RL-parsing}
Dani Yogatama, Phil Blunsom, Chris Dyer, Edward Grefenstette, and Wang Ling.
  2017.
\newblock Learning to compose words into sentences with reinforcement learning.
\newblock In \emph{International Conference on Learning Representations}.

\bibitem[{Zaremoodi and Haffari(2018)}]{zaremoodi-haffari-2018-incorporating}
Poorya Zaremoodi and Gholamreza Haffari. 2018.
\newblock \href {https://www.aclweb.org/anthology/C18-1120} {Incorporating
  syntactic uncertainty in neural machine translation with a forest-to-sequence
  model}.
\newblock In \emph{Proceedings of the 27th International Conference on
  Computational Linguistics}.

\end{thebibliography}
\bibliographystyle{acl_natbib}
\clearpage

\appendix
\section{Visualization of Self-Attention}\label{app:a}
In this section, we plot the heat maps of attention probability $E$ from the original Transformer and one from Tree Transformer.
In order to investigate whether models learn layer-wise tree structures, we average the scores of the attention heads within the same layer.

\begin{figure*}
\centering  
\subfigure[The layer 0.]{\includegraphics[width=0.495\linewidth]{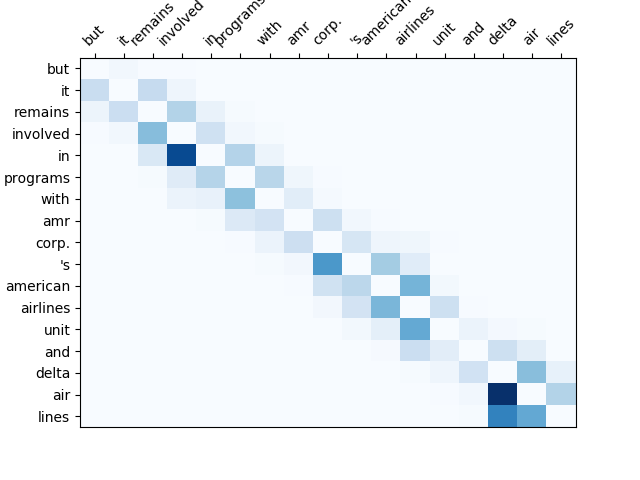}}
\subfigure[The layer 3.]{\includegraphics[width=0.495\linewidth]{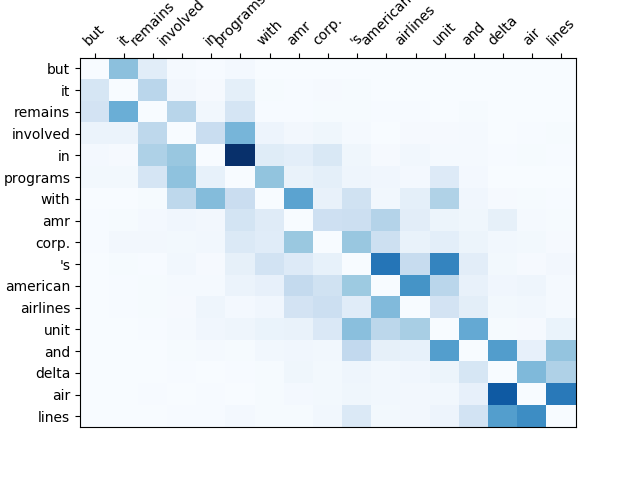}}
\subfigure[The layer 6.]{\includegraphics[width=0.495\linewidth]{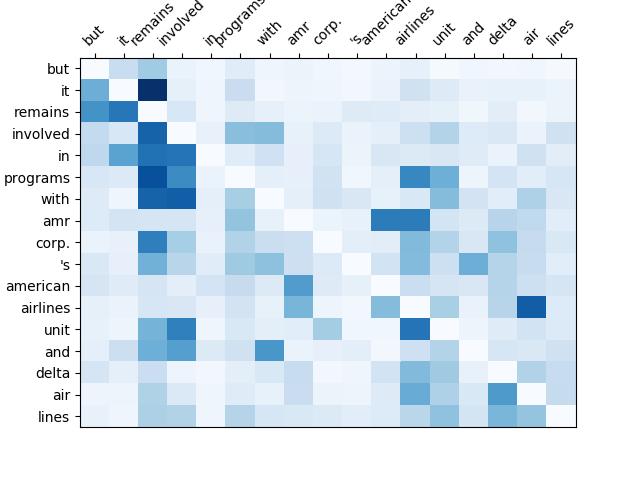}}
\subfigure[The layer 9.]{\includegraphics[width=0.495\linewidth]{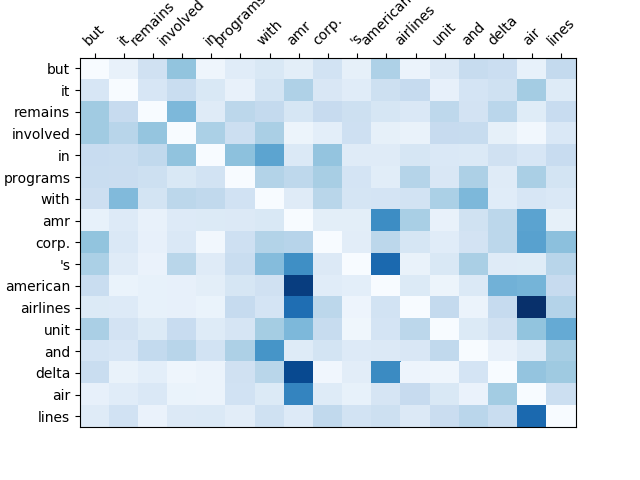}}
\caption{The attention heat map from the Tree Transformer.} 
  \label{fig:layer}
\end{figure*}

In the heat map from first layer of Tree Transformer (Figure~\ref{fig:layer}(a)), as the size of constituent in the layer is small, the words only attend to its adjacent words.
We also observe that the words within the same phrase attend to each other, such as the phrasal verb ``\emph{involve in}'' and the noun phrase ``\emph{delta air line}''.
With the constituents attaching to each other and becoming larger, in Figure~\ref{fig:layer}(b)-(d), the words attend to more distant words.

\begin{figure*}
\centering  
\subfigure[The layer 0.]{\includegraphics[width=0.495\linewidth]{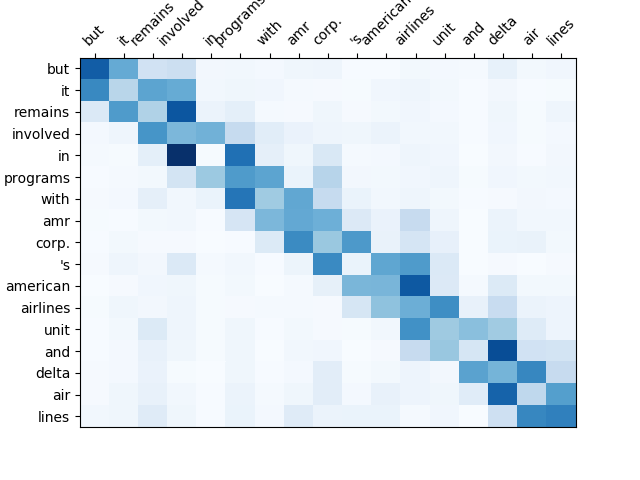}}
\subfigure[The layer 3.]{\includegraphics[width=0.495\linewidth]{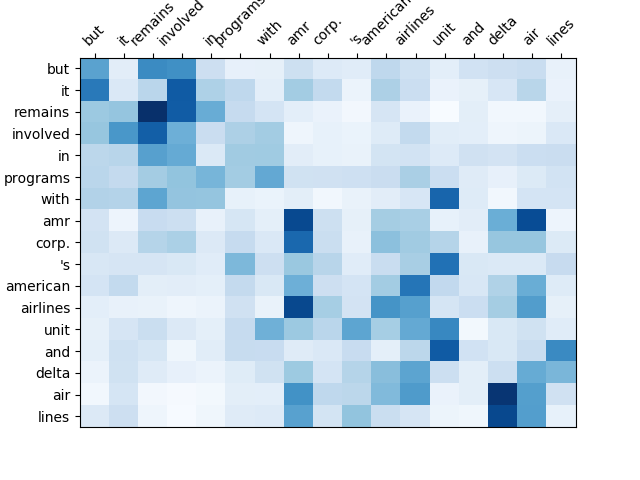}}
\caption{The attention heat map from the original Transformer.} 
  \label{fig:transformer_layer}
\end{figure*}

On the other hard, we average the scores of attention heads from each layer of the original Transformer and visualize them.
In the attention heat maps of the original Transformer, we also observe that it learns hierarchical structures in the first layer (Figure~\ref{fig:transformer_layer}(a)) that the words mostly attend to its neighboring words.
However, this phenomenon is only observed in the first layer that in the third layer (Figure~\ref{fig:transformer_layer}(b)), the attention heads attend to other words at will.

\clearpage

\iffalse
\begin{figure}[t]
  \centering
    \includegraphics[width=\linewidth]{layer_0.png}
      \caption{The attention heat map from the layer 0.} 
  \label{fig:layer0}
\end{figure}

\begin{figure}[t]
  \centering
    \includegraphics[width=\linewidth]{layer_3.png}
      \caption{The attention heat map from the layer 3.} 
  \label{fig:layer3}
\end{figure}

\begin{figure}[t]
  \centering
    \includegraphics[width=\linewidth]{layer_6.png}
      \caption{The attention heat map from the layer 6.} 
  \label{fig:layer6}
\end{figure}

\begin{figure}[t]
  \centering
    \includegraphics[width=\linewidth]{layer_9.png}
      \caption{The attention heat map from the layer 9.} 
  \label{fig:layer9}
\end{figure}

\begin{figure}[t]
  \centering
    \includegraphics[width=\linewidth]{transformer_layer_0.png}
      \caption{The attention heat map from original Transformer layer 0.} 
  \label{fig:transformer_layer0}
\end{figure}

\begin{figure}[t]
  \centering
    \includegraphics[width=\linewidth]{transformer_layer_3.png}
      
\end{figure}
\fi

\begin{figure*}[t!]
\centering  
\subfigure{\includegraphics[width=\linewidth]{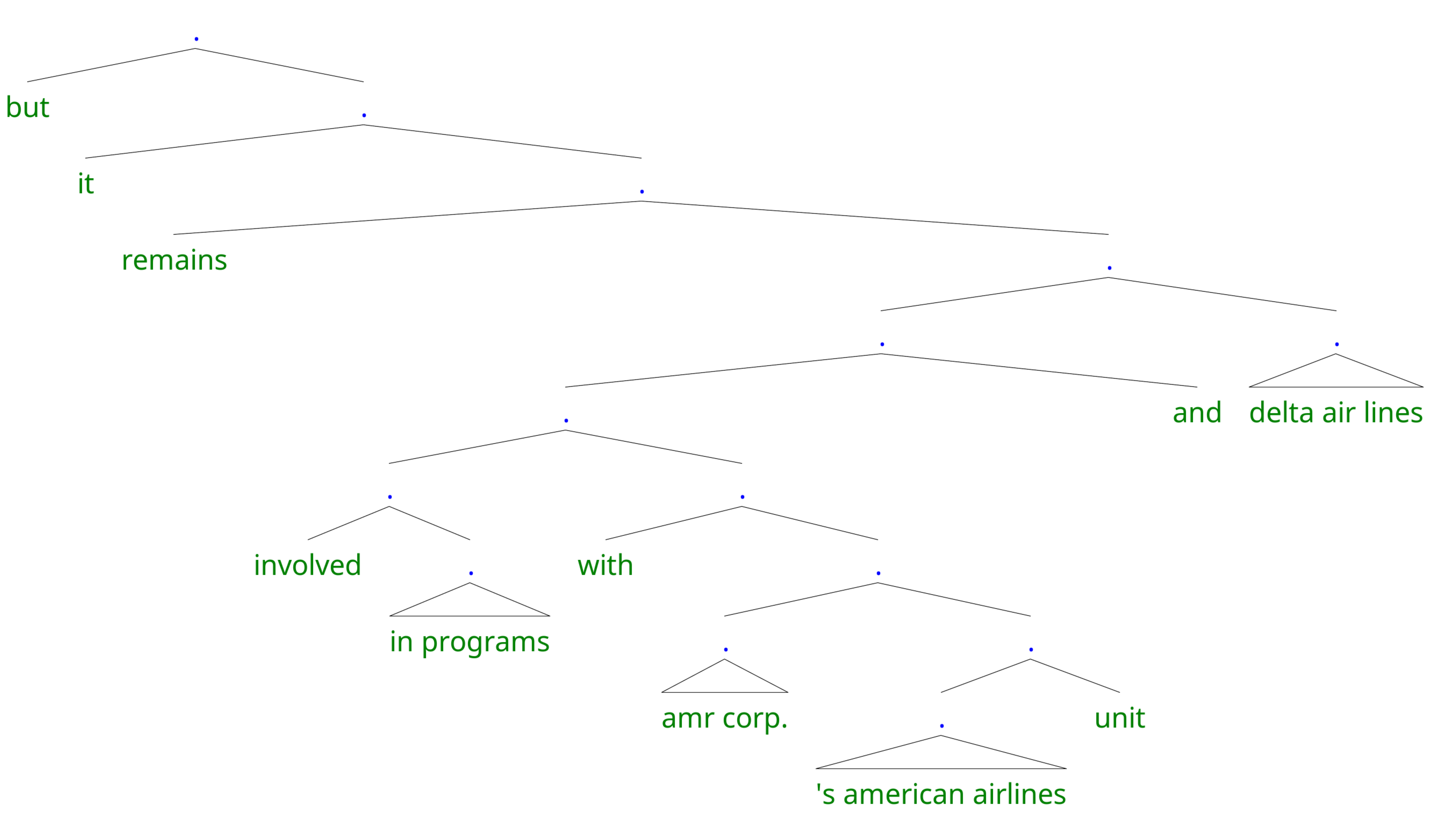}}
\subfigure{\includegraphics[width=\linewidth]{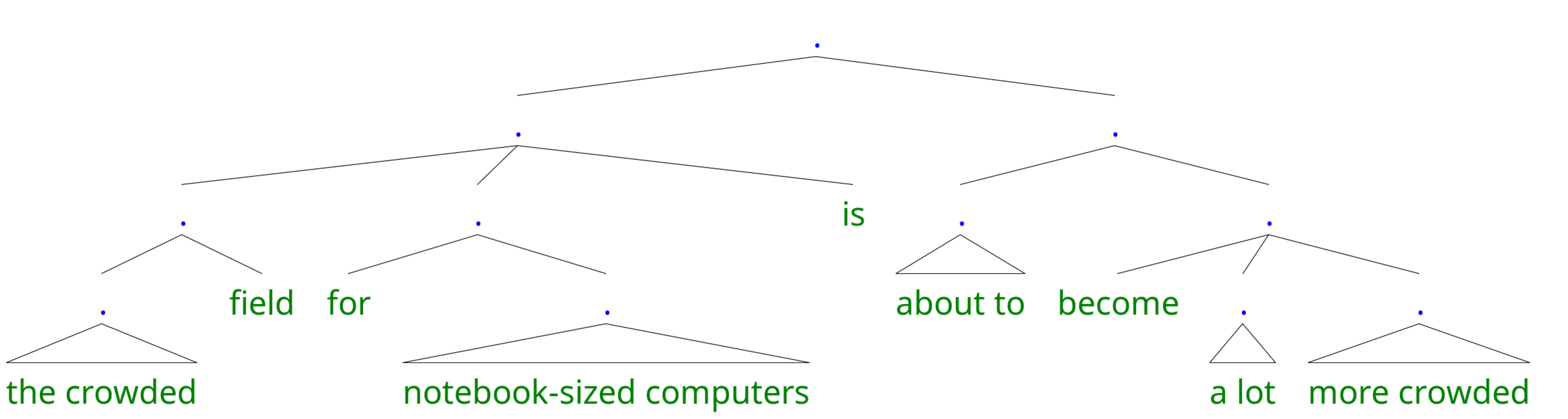}}
\subfigure{\includegraphics[width=.7\linewidth]{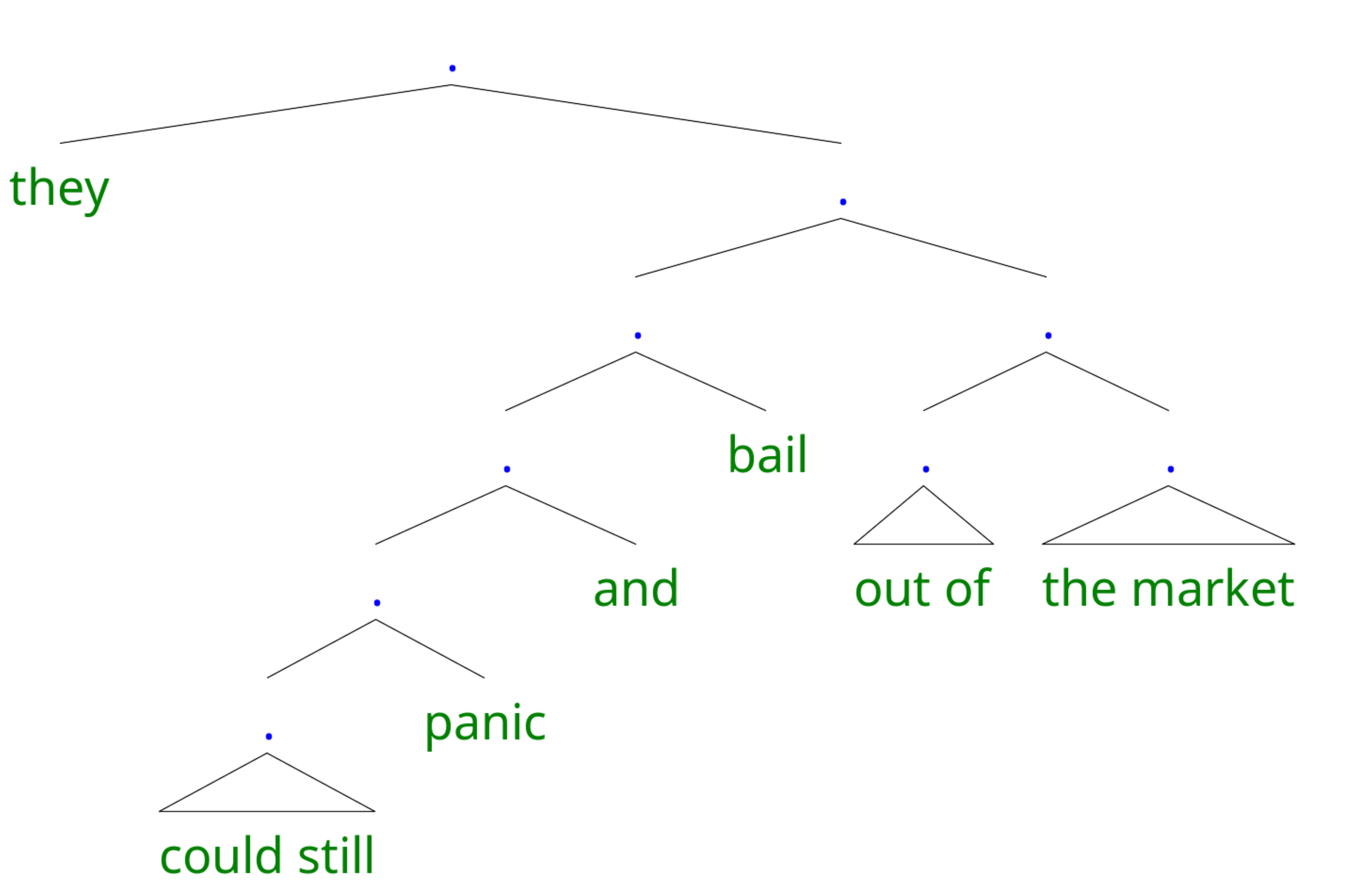}}
\caption{The parse trees induced by the proposed Tree Transformer.} 
  \label{fig:tree1}
\end{figure*}

\begin{figure*}[t!]
\centering  
\subfigure{\includegraphics[width=\linewidth]{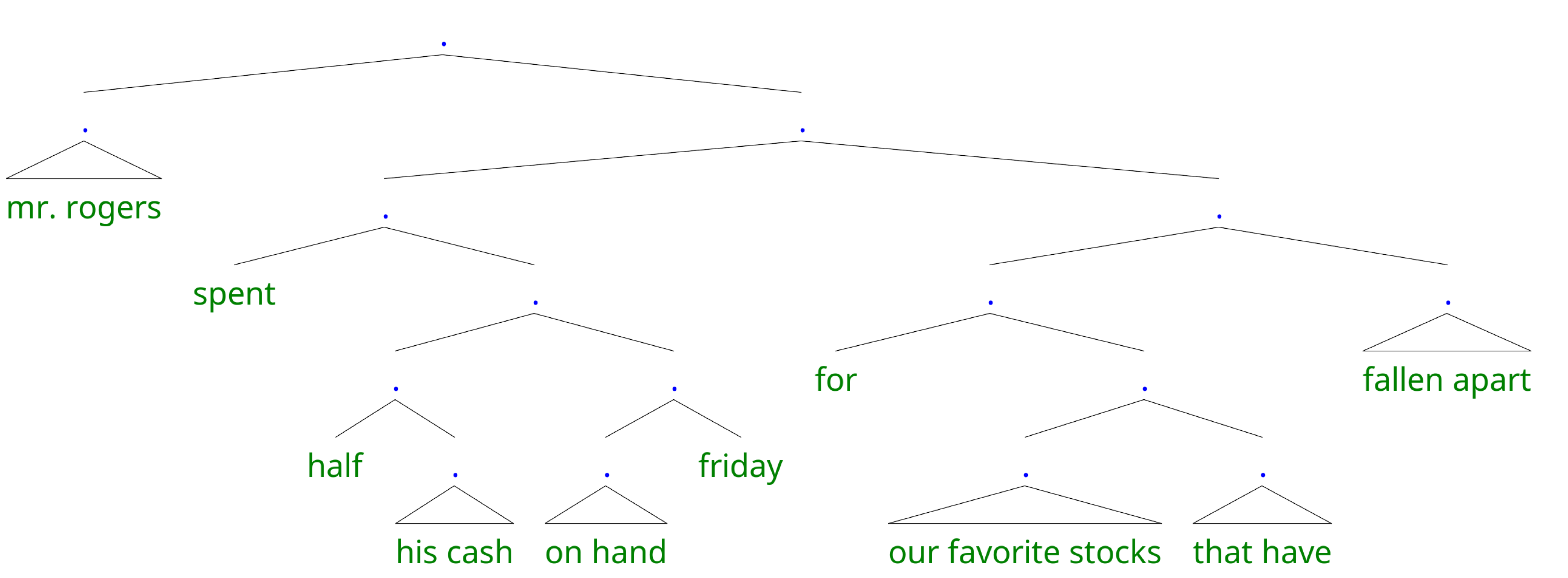}}
\subfigure{\includegraphics[width=.75\linewidth]{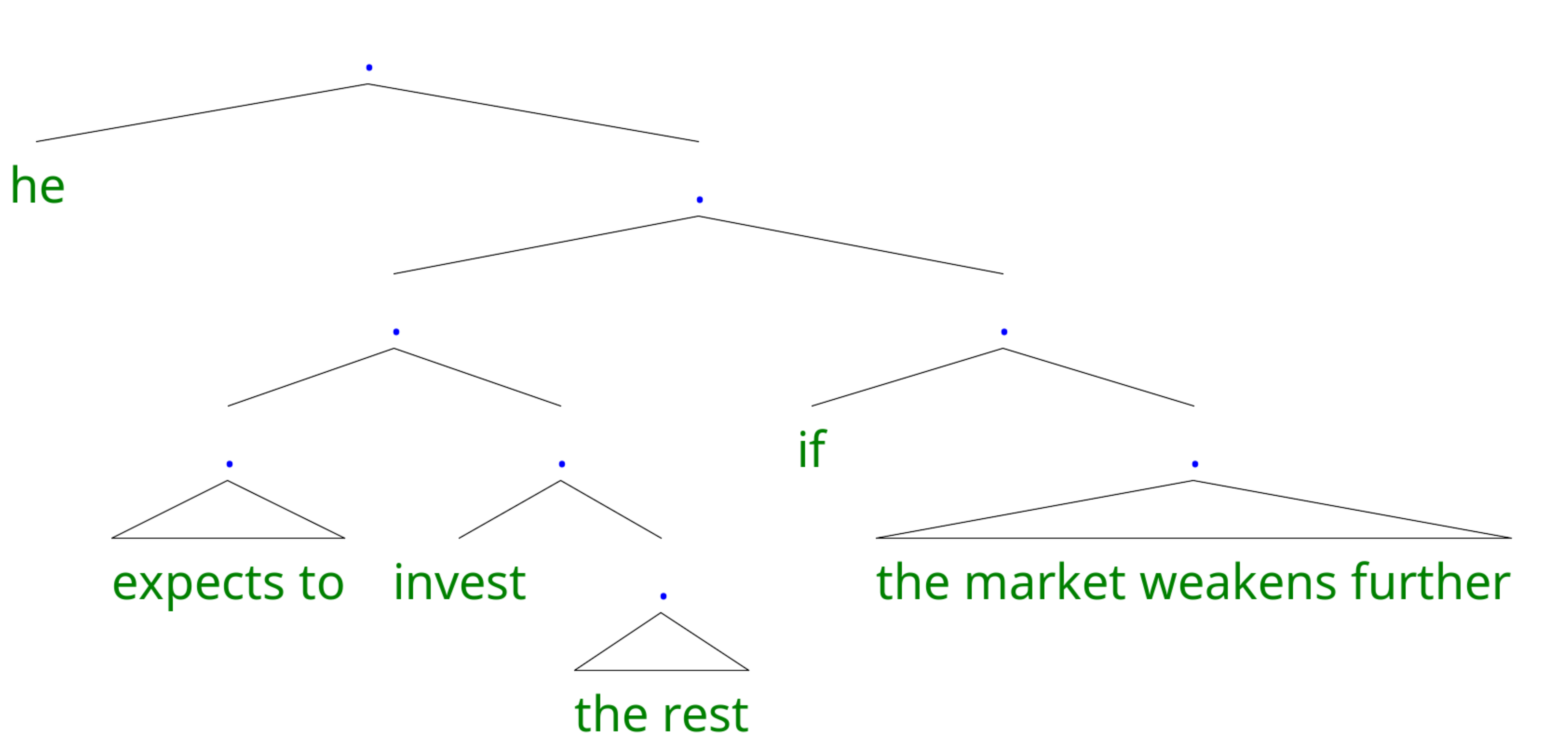}}
\subfigure{\includegraphics[width=\linewidth]{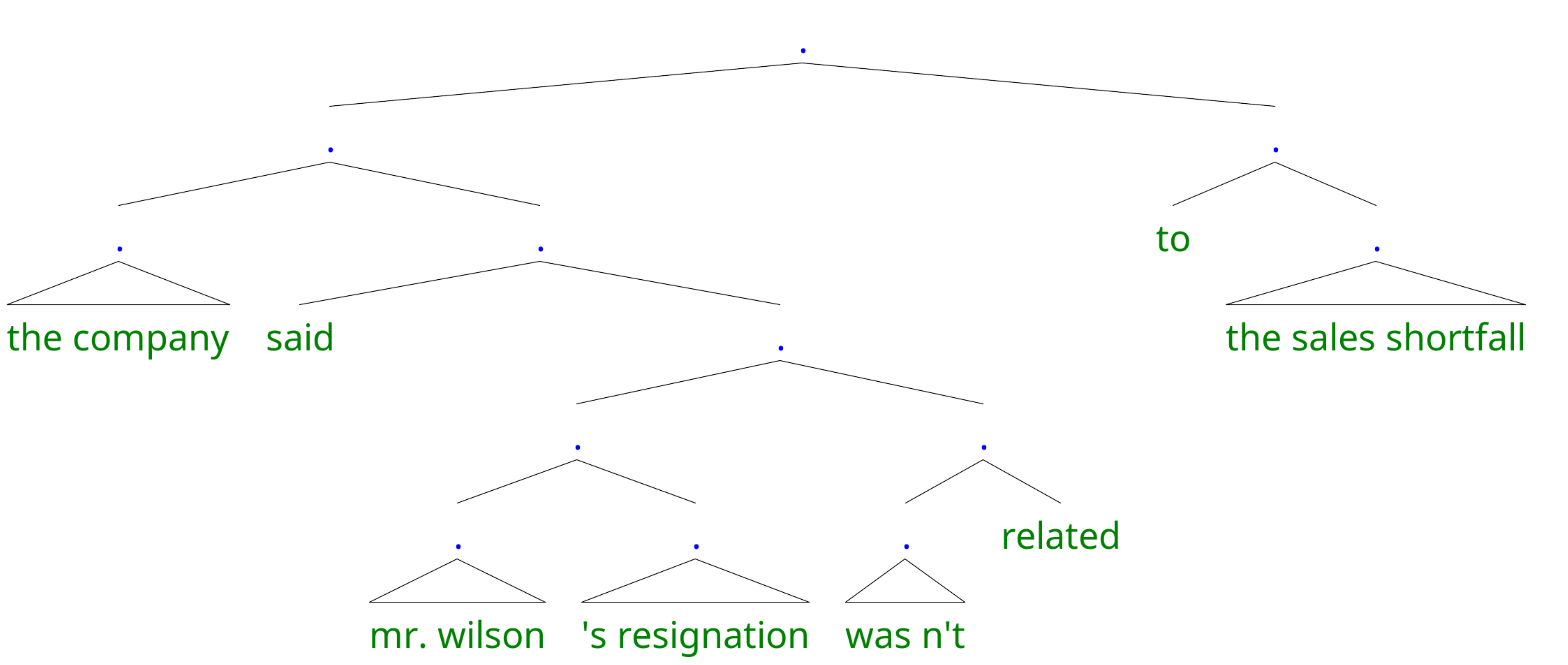}}
\caption{The parse trees induced by the proposed Tree Transformer.} 
  \label{fig:tree2}
\end{figure*}
 
\iffalse
\begin{figure*}[t]
  \centering
    \includegraphics[width=\linewidth]{tree0}
      \caption{A parse tree induced by Tree Transformer.}
  \label{fig:tree0}
\end{figure*}
%Lee: 這張圖上的字太小了

\begin{figure*}[t]
  \centering
    \includegraphics[width=\linewidth]{tree2}
      \caption{A parse tree induced by Tree Transformer.}
  \label{fig:tree2}
\end{figure*}

\begin{figure*}[t]
  \centering
    \includegraphics[width=120mm,scale=0.5]{tree3}
      \caption{A parse tree induced by Tree Transformer.}
  \label{fig:tree3}
\end{figure*}

\begin{figure*}[t]
  \centering
    \includegraphics[width=\linewidth]{tree4}
      \caption{A parse tree induced by Tree Transformer.}
  \label{fig:tree4}
\end{figure*}

\begin{figure*}[t]
  \centering
    \includegraphics[width=120mm,scale=0.5]{tree5}
      \caption{A parse tree induced by Tree Transformer.}
  \label{fig:tree5}
\end{figure*}

\begin{figure*}[t]
  \centering
    \includegraphics[width=\linewidth]{tree6}
      \caption{A parse tree induced by Tree Transformer.}
  \label{fig:tree6}
\end{figure*}
\fi

\end{document}